\documentclass[12pt]{article}
\usepackage[margin=1in]{geometry}

\usepackage[dvipsnames]{xcolor}
\usepackage[unicode=true]{hyperref}
\usepackage[english]{babel}
\usepackage{graphicx} 
\usepackage{amsmath}
\usepackage{amsthm}
\usepackage{mathtools}
\usepackage{amssymb}
\usepackage{authblk}
\usepackage{algpseudocode}
\usepackage{algorithm}
\usepackage{thmtools}
\usepackage{thm-restate}

\definecolor{ref_color}{rgb}{0, 0.47265625, 0.63671875}
\definecolor{mygreen}{rgb}{0,0.69,0.31}
\definecolor{myblue}{rgb}{0,0.44,0.75}
\definecolor{mypurple}{rgb}{0.44,0.19,0.63}

\newtheorem{assumption}{Assumption}
\newtheorem{hypothesis}{Hypothesis}
\newtheorem*{proof*}{Proof}

\hypersetup{
    colorlinks=true,
    linkcolor=ref_color,
    urlcolor=ref_color,
    }

\newcommand{\beginsupplement}{%
        \newpage
        \textbf{\huge{Supplementary Materials}}
        \vspace{1cm}
        \setcounter{section}{0}
        \renewcommand{\thesection}{S\arabic{section}}
        \renewcommand{\theHsection}{S\arabic{section}}
        \setcounter{table}{0}
        \renewcommand{\thetable}{S\arabic{table}}%
        \renewcommand{\theHtable}{S\arabic{table}}
        \setcounter{figure}{0}
        \renewcommand{\thefigure}{S\arabic{figure}}%
        \renewcommand{\theHfigure}{S\arabic{figure}}
     }

\title{Spiking Network Initialisation and Firing Rate Collapse}
\author[1]{Nicolas Perez-Nieves}
\author[1]{Dan F.M Goodman }
\affil[1]{Department of Electrical and Electronic Engineering, Imperial College London}
\date{}
\begin{document}

\maketitle

\begin{abstract}
   In recent years, newly developed methods to train spiking neural networks (SNNs) have rendered them as a plausible alternative to Artificial Neural Networks (ANNs) in terms of accuracy, while at the same time being much more energy efficient at inference and potentially at training time. However, it is still unclear what constitutes a good initialisation for an SNN. We often use initialisation schemes developed for ANN training which are often inadequate and require manual tuning. In this paper, we attempt to tackle this issue by using techniques from the ANN initialisation literature as well as computational neuroscience results. We show that the problem of weight initialisation for ANNs is a more nuanced problem than it is for ANNs due to the spike-and-reset non-linearity of SNNs and the firing rate collapse problem. We firstly identify and propose several solutions to the firing rate collapse problem under different sets of assumptions which successfully solve the issue by leveraging classical random walk and Wiener processes results. Secondly, we devise a general strategy for SNN initialisation which combines variance propagation techniques from ANNs and different methods to obtain the expected firing rate and membrane potential distribution based on diffusion and shot-noise approximations. Altogether, we obtain theoretical results to solve the SNN initialisation which consider the membrane potential distribution in the presence of a threshold. Yet, to what extent can these methods be successfully applied to SNNs on real datasets remains an open question.
\end{abstract}

\section{Introduction}


The remarkable success of deep artificial neural networks (ANNs) \cite{schmidhuber2015deep, lecun2015deep} has prompted a new interest in training spiking neural networks (SNNs) with the aim of studying information processing in the brain and advancing neuromorphic engineering \cite{richards2019deep, zenke2021visualizing}. ANNs are typically trained using a gradient descent algorithm that relies on the differentiability of the loss function with respect to the network parameters. However, due to the binary nature of spiking neurons activations, this assumption does not hold true in SNNs. Different methods have been developed to circumvent this issue, one of the most popular being surrogate gradient descent \cite{neftci2019surrogate, zenke2021remarkable}. Surrogate gradient algorithms use a smooth function to approximate the spiking function (Heaviside step function) in the gradient computation. This means that the network still propagates spikes in the forward pass but backpropagates continuous values. Since there are many possible choices of the surrogate function the surrogate gradient is, unlike the true gradient, not unique. Yet, SNNs have been shown to learn very reliably under a wide variety of surrogate choices provided the network is properly initialised \cite{zenke2021remarkable}. 

Similarly to ANNs, a suboptimal weight initialisation on SNNs may result in vanishing or exploding gradients \cite{hochreiter1997long, rossbroich2022fluctuation}. This issue has been largely explored in the ANN literature and a wide range of strategies has been developed for an optimal parameter initialisation \cite{skorski2021revisiting}. The most popular methods rely on variance flow analysis and aim to keep constant activations and gradient variance across layers. Depending on the activation function and weight distribution different initialisation schemes have been devised \cite{glorot2010understanding, he2015delving, xu2016revise, arpit2019initialize}. Other methods have also been developed based on mean-field theory \cite{xiao2018dynamical}, Lipschitz property \cite{virmaux2018lipschitz} and Hessian approximations \cite{skorski2021revisiting} (see \cite{skorski2021revisiting} for a review). 

Spiking Neural Networks are currently initialised following schemes designed for ANNs. These schemes are inadequate as the SNN activation function is different from the ones studied in the ANN literature and, unlike SNNs, ANNs do not have any temporal dynamics or resetting. Typically, this leads to either having too few spikes in the final layers (vanishing activity) or having saturated neurons that fire synchronously at their maximum rate and cannot propagate any information (saturated/exploding activity). As a consequence, we often end up having to manually adjust the weight variance and need to use regularisation penalties to ensure appropriate rates. This problem becomes worse when considering deeper networks where potentially each layer needs to be tuned separately as the spiking signals get lost or gradients vanish. 

Research on optimal initialisation schemes on SNNs are very scarce. To the best of our knowledge, while different initialisations have been used in the literature \cite{bellec2018long, zenke2021remarkable, herranz2022surrogate}, the problem of SNN initialisation has only been directly addressed in \cite{ding2022accelerating} and \cite{rossbroich2022fluctuation}. However, these works presents important limitations. Firstly, in \cite{ding2022accelerating} they consider the asymptotic spiking response given a mean-driven input (i.e. low noise). This limits the possibility of running the SNN in the most energy efficient regime with as few spikes as possible. Secondly, \cite{rossbroich2022fluctuation} only considers approximate input-output curves and membrane distributions that do not include the spiking and resetting mechanisms which ultimately limit the applicability of this method to deeper networks. Finally, both methods concentrate most of the analysis on the forward pass and do not derive analytical results for the backward pass. In this paper, we aim to extend these works by considering:

\begin{itemize}
    \item \textbf{Variance propagation in forward and backward pass}. We use methods developed for ANNs on SNNs forward and backward propagation. This leads to obtaining a weak bound for the firing rate in the forward pass and a closed form expression for the optimal weight variance in the backward pass.
    \item \textbf{Fokker-Planck and Shot-noise approach for the forward pass}. Since variance propagation is not enough to establish the firing rate in the final layer we resort to the diffusion approximation and similar approaches to compute the output firing rate.
    \item \textbf{Firing rate collapse solution}. As we increase the timestep to typical step sizes used to train SNNs with surrogate gradient methods, the diffusion approximation becomes less accurate. We introduce three different solutions to this problem under different assumptions.
\end{itemize}

\section{Background Theory} \label{sec:05_02_background}

In this section we present the main theoretical results from which we begin our derivations. These include the basic spiking neural network model, forward and backward variance propagation on ANNs for optimal weight initialisation and different methods to compute the output firing rate and membrane distribution of a layer of LIF neurons including: diffusion approximation, shot-noise and threshold integration. We only state the main results here. For a more through derivation of them we refer to the Supplementary Materials \ref{supp_sec:05_02_background}. 

\subsection{Spiking Neural Networks}

We consider networks of Leaky Integrate and Fire (LIF) neurons that receive stochastic spikes from delta synapses. Neurons are organised in a total of $L$ layers with $n^{(l)}$ neurons each for $l=0, \ldots, L-1$. The membrane potential $v_j^{(l)}(t)$ of the $jth$ neuron in layer $l$ evolves according to

\begin{align}\label{eq:02_02_lif_mem_rest}
    \tau \frac{d v_j^{(l)}(t)}{dt} = - (v_j^{(l)}(t)+ R I_{ext}) + \sum_{i}^{n^{(l-1)}} \sum_{t_i^{(f)}}w_{ij}^{(l-1)} \delta(t-t_i^{(f)}), \qquad \text{if } v_j^{(l)}(t)<V_{th}
\end{align}

According to this equation, the membrane integrates incoming spikes coming from a total of $n^{(l-1)}$ sources.  These sources can be either the neurons of the previous layer or in the case of the first layer some spike sources (e.g. Poisson generators). Incoming spikes are weighted by $w_{ij}\in \mathbb{R}$. Upon reaching the threshold $V_{th}$ three things happen: a formal event called \textit{spike} is emitted, the membrane is reset to $V_{r}$ and the neuron enters a refractory period that lasts $t_{ref}$ seconds where the neuron cannot spike and the membrane potential is kept constant at $V_r$. We will assume $R=1$ and $t_{ref}=0$ unless otherwise specified.

We can write the output spikes as a \textit{spike train} given by 

\begin{equation}\label{eq:02_02_lif_spk}
    s_j^{(l)}(t) = \sum_{t_j^{(f)}} \delta(t-t_j^{(f)})
\end{equation}

where we have omitted the superscript $l$ in the times $t_i^{(f)}$ for notational simplicity.

We are interested in quantifying the level of activity of a layer. We can use the \textit{population activity} for this purpose. This can be defined as the proportion of neurons at layer $l$ that fire within a short time interval $[t, t+\Delta t]$. 

\begin{equation}
    \nu^{(l)}(t) = \frac{1}{n^{(l)}} \sum_{i=1}^{n^{(l)}} \sum_{t_j^{(f)}} \delta (t-t_j^{(f)}), \qquad t_j^{(f)} \in [t, t+\Delta t]
\end{equation}

\subsection{Population Firing rate in Spiking Neural Networks} \label{sec:05_02_diff_shot}

Computing the firing rate of a spiking neural network is far from trivial due to the one-sided threshold-and-reset mechanism of spiking neurons. Several approaches have been developed to obtain it under different assumptions. These include the diffusion approximation \cite{gerstner2014neuronal}, which assumes that the weights are small compared to the distance between the reset and threshold and that weights are restricted to a total of $K$ different values which must include both positive and negative values. We can allow for larger weights as long as they are distributed following exponential distributions. This is called shot-noise and generalises the diffusion approximation \cite{richardson2010firing}. Both of these methods are restricted to LIF neurons. For other neuron models the threshold integration method can be used \cite{richardson2007firing, richardson2008spike}. In addition to the particular assumption to each method, all of them assume a homogeneous (identical neurons) population of neurons and the input spikes being generated by a finite number of homogeneous Poisson processes. We refer to Supplementary Materials \ref{supp_sec:05_02_background} for more details on these methods. 

The diffusion and shot-noise methods allow us to obtain closed form solutions to compute the theoretical output firing rate $r_{out}$ of a population of neurons. These are given by

\begin{align}
\frac{1}{r_{out}}= \tau \sqrt{\pi} \int_{\frac{V_r-\mu}{\sigma}}^{\frac{V_{th}-\mu}{\sigma}} \exp{\left( x^2\right)}\left( 1 + \text{erf}(x) \right) dx\
\label{eq:05_02_siegert}
\end{align}

for the diffusion case, with $\mu\!=\!R I_{ext} + \tau \sum_k r_k w_k$ and $\sigma ^2 \!=\!\tau \sum_k r_k w_k^2$ given $K$ synaptic input types with weights $w_k$ and rates $r_k$.

For the shot-noise case we have

\begin{align}
    \frac{1}{r_{out}} = \tau \int_0^{1/w_e}\frac{Z^{-1}_0(x)}{x} \left( \frac{e^{xV_{th}}}{1-xw_e}-e^{xV_r} \right) dx
    \label{eq:05_02_shot_noise_rate}
\end{align}

with $Z_0^{-1}(x) = (1-xw_e)^{\tau r_e}(1-xw_i)^{\tau r_i}$ where $r_e$ and $r_i$ are the inhibitory and excitatory firing rates respectively and weights are drawn from exponential distributions with parameters $w_e$ and $w_i$ respectively.

\subsection{Initialisation based on variance flow in ANNs} \label{subsec:05_02_var_prop}

The results in this subsection are are taken from \cite{he2015delving}. Unlike SNNs, ANNs do not have any dynamics or resetting present, they simply consist of an affine transformation followed by a non-linearity.

\begin{align}
    y^{(l+1)} &= W^{(l)}x^{(l)} + b^{(l)} \\
    x^{(l+1)} &= f(y^{(l+1)})
\end{align}

where $y^{(l+1)}, x^{(l+1)} \in \mathbb{R}^{n^{(l+1)}}$, $W \in \mathbb{R}^{{n^{(l+1)}}\times {n^{(l)}}}$, $x^{(l)} \in \mathbb{R}^{n^{(l)}}$, $b^{(l)}=\pmb{0}$ at initialisation, $f$ is a non-linear element-wise operation and $l=0, \ldots, L\!-\!1$. We make the assumptions that $x_i^{(l)}$ are i.i.d. for all $i$ and $l$, $W^{(l)}_{ij}$ and $x^{(l)}_{i}$ are independent and $W^{(l)}_{ij}$ is zero mean and symmetric around zero and i.i.d for all $i$, $j$ and $l$. We also assume ReLU activations. Then it can be shown that for $L$ layers

\begin{align}
    Var\left[y_j^{(L)} \right]&= Var\left[  y_i^{(1)} \right] \prod_{l=1}^{L-1} \frac{1}{2} n^{(l)}Var\left[  W^{(l)}_{ij} \right]
\end{align}

Ensuring that the variance of the activations does not increase or decrease exponentially with the number of layers, results in Kaiming's initialisation \cite{he2015delving}

\begin{align} \label{eq:05_02_ann_init_forward}
    Var\left[ W^{(l)}_{ij} \right] = 2/n^{(l)}
\end{align}

A similar result can be derived for the backward pass. Given the gradients

\begin{align}
    \Delta x^{(l)} &= \Delta y^{(l+1)} W^{(l)} \\
    \Delta y_i^{(l+1)} &= \Delta x_i^{(l+1)} f'\left(y_i^{(l+1)}\right)
\end{align}

where we use $\frac{\partial \varepsilon}{\partial z }\!:=\!\Delta z$ for any partial derivative of the loss with respect to $z$ as in \cite{he2015delving} with $\varepsilon: \mathbb{R}^{n^{(L\!-\!1)}} \to \mathbb{R}$ being an arbitrary loss function. As before, for $L$ layers it can be shown that

\begin{align}
    Var\left[\Delta x_j^{(1)} \right]&= Var\left[ \Delta x_i^{(L-1)} \right] \prod_{l=1}^{L-2} \frac{1}{2} n^{(l+1)}Var\left[  W^{(l)}_{ij} \right]
\end{align}

and to avoid exponential variance growth/shrink we set

\begin{align} \label{eq:05_02_ann_init_backward}
    Var\left[ W^{(l)}_{ij} \right] = 2/n^{(l+1)}
\end{align}

The expressions found for backward and forward optimal initialisation are different \eqref{eq:05_02_ann_init_forward},\eqref{eq:05_02_ann_init_backward}. However, choosing any of them leads to the forward(backward) variance being $1$ while the backward(forward) variance is still a constant \cite{he2015delving}.
\section{Firing Rate Collapse}

In principle, we should be able to obtain the expected firing rate with any of the three methods mentioned in Section \ref{sec:05_02_background}. However, these methods assume a continuous time neuron model when in fact neurons are simulated in discrete time. Thus, as we increase to simulation timestep we find that these predictions worsen. 

This is hardly a problem in most of the computational neuroscience literature, as often it is perfectly feasible to simply decrease the simulation timestep. Nevertheless, as we start to deal with SNNs for deep learning the increasing memory and computational cost constrains the timestep to be about one order of magnitude larger (at around $1$ms) than in the computational neuroscience literature (usually below $0.1$ms). This change in temporal resolution is important enough that the effect of a larger timestep can no longer be ignored.

The following sections start by illustrating the problem and several hypotheses as to why it happens. Then, we propose three different types of solutions for it under different assumptions.

\subsection{Firing Rate Collapse problem identification}

The continuous time expression for the subthreshold membrane potential under a constant bias current and random input spikes after integrating the input for $\Delta t$ seconds is obtained by solving \eqref{eq:02_02_lif_mem_rest} and is given by

\begin{align}\label{eq:05_03_lif_mem_sol_syn} 
    v(t) &= \underbrace{\alpha v(t-\Delta t) + (1-\alpha)RI}_{\text{Deterministic}} + \underbrace{\sum_i \sum_{t_{i}^{(f)}}w_{i}\exp\left(-\frac{t -t_i^{(f)}}{\tau_m}\right)H(t-t_i^{(f)})}_{\text{Stochastic}}, \quad v(t)<V_{th} 
\end{align}

with $\alpha\!=\!\exp(-\Delta t / \tau_m)$ and $t_{i}^{(f)} \in [t, t\!+\!\Delta t]$.

Note that the spike integration in \eqref{eq:05_03_lif_mem_sol_syn} requires evaluating an exponential at continuous times $t\!-\! t_i^{(f)}, \: \forall i$. This is problematic if we want to use a discrete uniform time step $\Delta t$ as all spikes are synchronised to a grid with $\Delta t$ resolution. Thus, we cannot generate more than one spike during a time interval either on the input or the output.

\begin{figure}[!ht]
\centering
\includegraphics[width=0.8\textwidth]{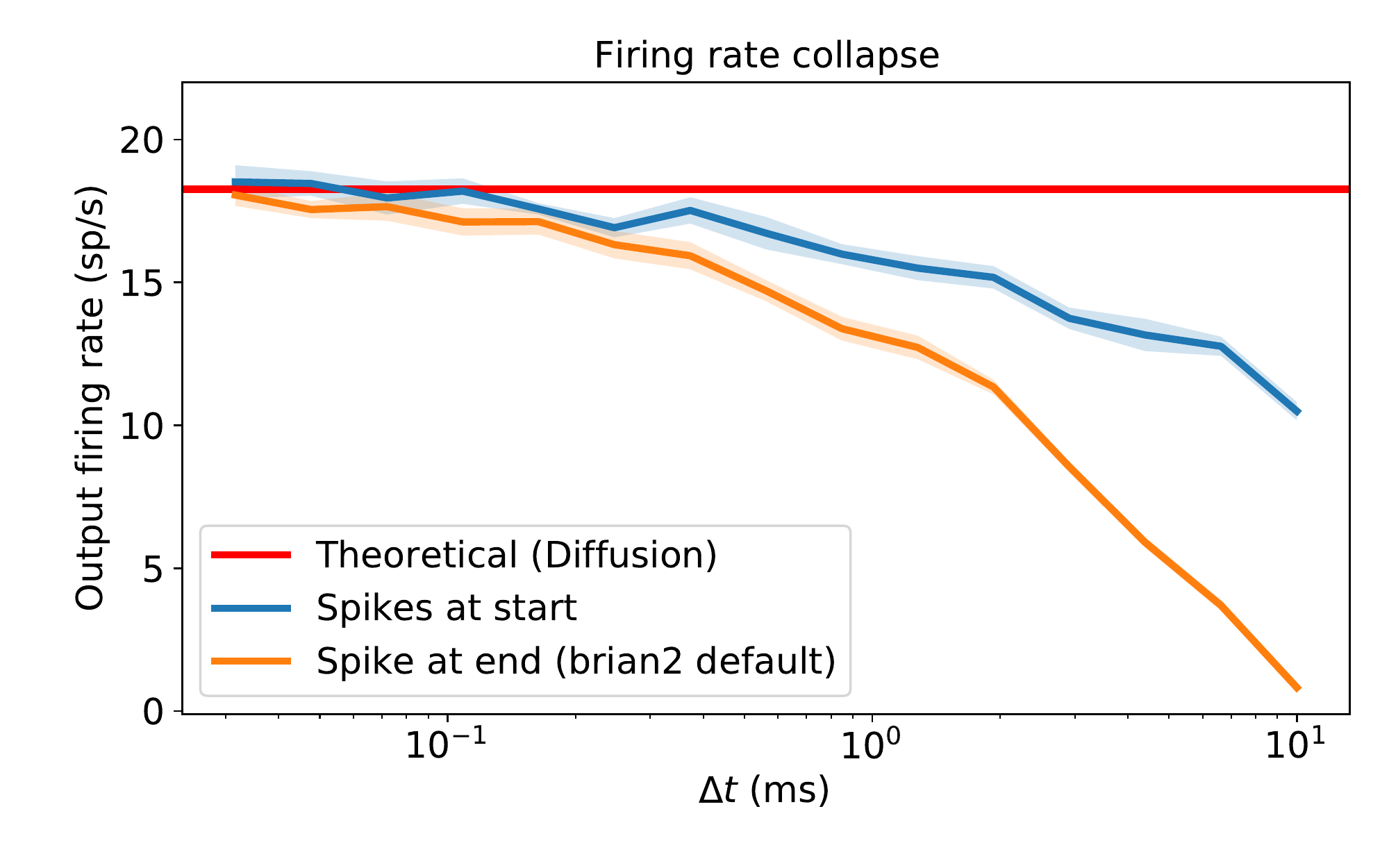}
\caption{Firing rate collapse as the simulation timestep increases. Average of $10$ simulations with standard error in the mean. Population of $1000$ LIF neurons receiving inputs from $1000$ Poisson inputs at $50$Hz each connected with a probability of $0.5$, the inhibitory and excitatory weights were $w_e\!=\!-w_i\!=\!0.01$. Other parameters were set to $\tau\!=\!10$ms, $I_{ext}\!=\!0.8$, $R\!=\!1$, $V_r\!=\!0$ and $V_{th}\!=\!1$.}
\label{fig:05_03_collapse_brian2}
\end{figure}

For this reason, most SNN simulators assume that all input spikes take place at a fixed point during the interval, often either at the beginning or at the end. The first thing we note is that if we assume that spikes are at the end, the exponential in \eqref{eq:05_03_lif_mem_sol_syn} tends to $0$ for large $\Delta t$. This is indeed the case for the default settings of some spiking neuron libraries like \texttt{brian2} \cite{Stimberg2019}. Figure \ref{fig:05_03_collapse_brian2} shows that the firing rate tends to zero as we increase $\Delta t$ as the effect of the input spikes becomes negligible.

\begin{figure}[!ht]
\centering
\includegraphics[width=\textwidth]{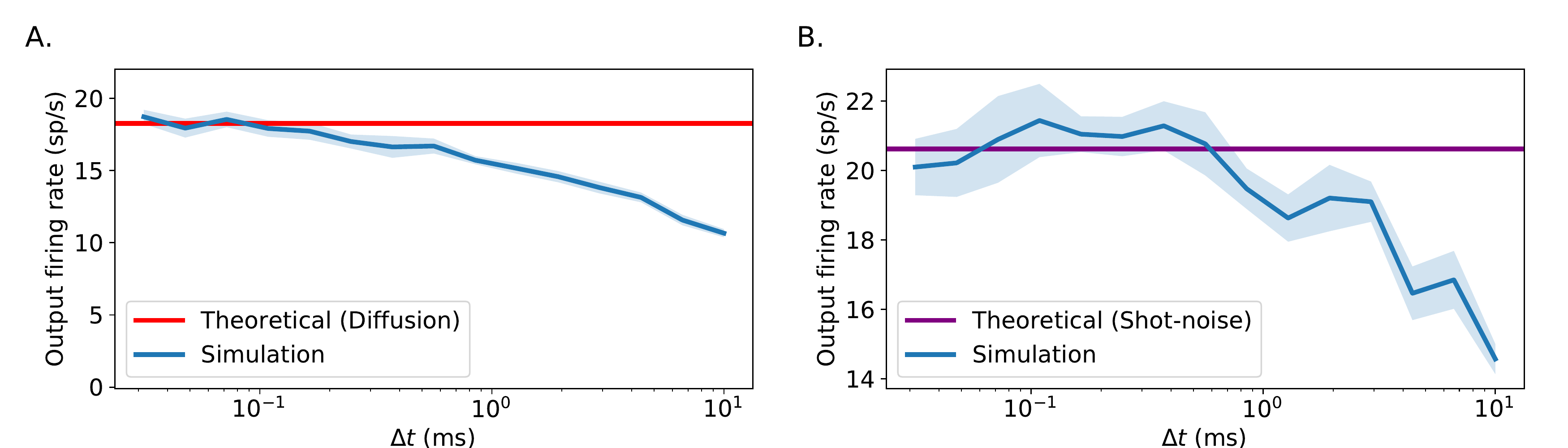}
\caption{Firing rate collapse as the simulation timestep increases. Average of $10$ simulations with standard error in the mean. \textbf{A.} Diffusion case. A population of $1000$ LIF neurons receiving inputs from $1000$ Poisson inputs at $50$Hz connected with a probability of $0.5$, the inhibitory and excitatory weights were $w_e\!=\!-w_i\!=\!0.01$. Other parameters were set to $\tau\!=\!10$ms, $I_{ext}\!=\!0.8$, $R\!=\!1$, $V_r\!=\!0$ and $V_{th}\!=\!1$. \textbf{B.} Shot-noise case with $3000$ output neurons and $1600$ Poisson sources at $0.14$Hz connected with probability $1$. Other parameters were set to $\tau\!=\!10$ms, $I_{ext}\!=\!0.$, $R\!=\!1$, $V_r\!=\!0$, $V_{th}\!=\!1$ and $a_e=-a_i\!=\!0.4$.}
\label{fig:05_03_collapse}
\end{figure}

Alternatively, we would expect that choosing the beginning of the interval would lead to a simulation firing rate closer to the theoretical rate, if not slightly higher, as the input spike strength has not decayed at all. Yet, this is not what we observe in Figure \ref{fig:05_03_collapse}. Instead, the firing rate still decays with the simulation timestep in both the diffusion and the shot noise case. Importantly, the firing rate collapse is already substantial at around $1$ms, thus challenging any predictions that we would make for deep SNN training (particularly so on the final layers).

\subsection{Hypothesis}

We can propose three different hypothesis to explain why the firing rate decays as we increase the timestep length:

\begin{hypothesis} \label{hyp:05_03_hyp1}
We cannot generate high enough input rates at large $\Delta t $ because each Poisson generator can generate at most a single spike per step.
\end{hypothesis}
\begin{hypothesis} \label{hyp:05_03_hyp2}
We cannot generate high enough output rates at large $\Delta t $ because each output neuron can only spike once per step.
\end{hypothesis}
\begin{hypothesis} \label{hyp:05_03_hyp3}
We are adding all inhibitory and excitatory spikes together at once but some permutation of the incoming spikes could lead to crossing the threshold before integrating the rest of the spikes.
\end{hypothesis}

Hypothesis \ref{hyp:05_03_hyp1} is a problem that only takes place if the rates are too high relative to the step length. For instance, looking at \ref{fig:05_03_collapse}A at $\Delta t\!=\!1 ms$ and given that input units fire at $50$Hz, they only fire more than once per timestep $0.12\%$ of the time. Thus, we should correct the input rate to $49.94$Hz. This however leads to a theoretical output rate of $18.25$Hz instead of $18.26$Hz when we empirically get about $16$Hz in the simulation. Thus, this hypothesis does not explain why we see such a drastic firing rate collapse.

Hypothesis \ref{hyp:05_03_hyp2} is similarly invalidated if we assume that the output neuron spike generation is also Poisson and note that the theoretical output rate of $18.26$Hz is lower than that of the input leading to spike more than once per interval just $0.016\%$ of the time.

\begin{figure}[!ht]
\centering
\includegraphics[width=0.9\textwidth]{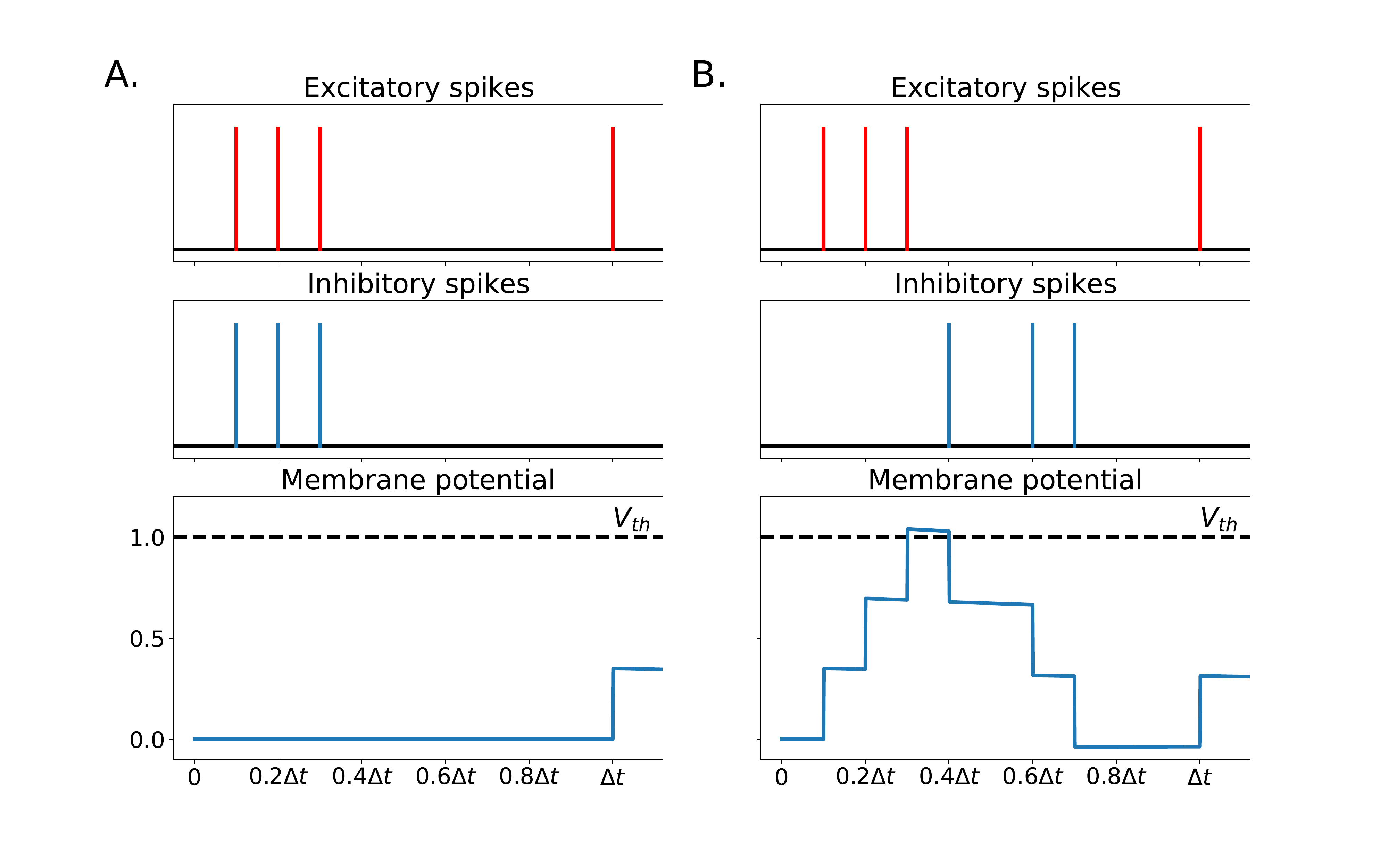}
\caption{Two different input spike orderings that lead to different results. In panel A we do not spike while in panel B we do. Yet, in both cases the net input is the same $w\left(N\!-\!M\right)\!=\!w(4\!-\!3)=w$. Integrating all spikes at once always leads to the first case.}
\label{fig:05_03_ordering}
\end{figure}

For the final hypothesis \ref{hyp:05_03_hyp3}, the key idea is that simulating all input spikes as arriving at the same time is equivalent to constraining the order of spike arrival by only considering orderings in which all excitatory inputs are perfectly aligned with an inhibitory input and thus only the difference of excitatory and inhibitory inputs is integrated as illustrated in Figure \ref{fig:05_03_ordering}A. Yet, other orderings which lead to spiking while  ending in the same final value are still possible as shown in Figure \ref{fig:05_03_ordering}B.

\subsection{General assumptions and proposed solutions}

For simplicity, since all input spikes come from independent Poisson sources we can merge them to get a single Poisson process with rate equal to the sum of the rates of each process. Then, the \emph{actual} value of membrane potential during a time step is given by

\begin{align} \label{eq:05_03_lif_correct}
    v(t+s\Delta t) = \alpha(s) v(t) + (1-\alpha(s))I +  \sum_{i=1}^{n}w_i \exp \left( -\frac{t+s\Delta t-t_i}{\tau} \right)H(t-t_i), \quad s\in [0, 1]
\end{align}

where $\alpha(s) = \exp(-s\Delta t / \tau)$, $n$ is the total number of spikes received during the interval and $t_i$ is a sequence of independent uniform random variables in $[t, t+\Delta t]$.

In order to simplify our analysis we will adopt the following assumptions:

\begin{assumption} \label{as:05_03_weight}
The spikes integrated during a timestep are not affected by the decay within that timestep.
\end{assumption}

\begin{assumption} \label{as:05_03_deterministic}
The deterministic part of the membrane potential during a timestep remains constant during that timestep and takes the value $\alpha v(t) + (1-\alpha)I$.
\end{assumption}

With these assumptions equation \eqref{eq:05_03_lif_correct} becomes 

\begin{align} \label{eq:05_03_lif_simplified}
    v(t+s\Delta t) = \alpha v(t) + (1-\alpha)I +  \sum_{1\leq i\leq \lfloor sn \rfloor}w_i, \quad s\in [0, 1]
\end{align}

Note that as a consequence of Assumption \ref{as:05_03_weight} the decaying exponentials multiplying the weights have been removed altogether and we are left with a piece-wise constant function where the input spikes are all equally spaced within the timestep. This may seem counter intuitive as it looks like we have lost some important timing information. However, as per Hypothesis \ref{hyp:05_03_hyp3}, the key reason why we loose output spikes is due to ordering of the spikes and Equation \eqref{eq:05_03_lif_simplified} still allows for all possible permutations. 

One reasonable alternative to equation \eqref{eq:05_03_lif_simplified} would be to use the average exponential value $\bar{\alpha}\!=\!\mathbb{E}\left(\exp(-t_i/\tau)\right)\!=\!\frac{\tau}{\Delta t}\left( 1-\exp(-\Delta t /\tau) \right)$ instead of $\alpha$ leading to an effective weight of $w_i\bar{\alpha}$. Note however that $\bar{\alpha}\to 0$ as we increase $\Delta t$ and thus we could end up in a similar situation as in Figure \ref{fig:05_03_collapse_brian2}. Besides, often, threshold crossing will take place for permutations with several excitatory spikes close together, thus the effective weight will be closer to $w_i$, leading again to Assumption \ref{as:05_03_weight}. Empirical results also showed that choosing $w_i\bar{\alpha}$ resulted in worse performance.

Similarly, we could have used the average decay value $\bar{\alpha}$ for the deterministic part. We choose to use $\alpha$ because the empirical results are slightly better when using this value as we will show in the following sections.

In the following three sections we analyse this problem under three different particular assumptions:

\begin{itemize}
    \itemsep0em
    \item \textbf{Two weights of the same magnitude and opposite sign:} We model the incoming spikes as a random walk which allows us to analytically compute the probability of firing.
    \item\textbf{Random weights and high number of spikes per interval:} We model the incoming spikes as a Wiener process which also allows us to analytically compute the probability of firing.
    \item \textbf{General case:} In the general case we have random weights from an arbitrary distribution and we do not constrain the spiking rate. We cannot derive an analytical solution for this case but we can devise an algorithm to simulate the correct behaviour by randomly choosing a permutation of the input spikes.
\end{itemize}


\subsection{Input spikes as a random walk}

\subsubsection{Approximating spike integration as a random walk}

We now assume that we only have two possible weight values corresponding to excitatory and inhibitory with the same magnitude, $w_{exc}\!=\!-w_{inh}\!=\!w$. Then, we can write the membrane potential within that timestep as

\begin{align} \label{eq:05_03_lif_random_walk}
    v(t+s\Delta t) = \alpha v(t) + (1-\alpha)I +  w\sum_{1\leq i\leq \lfloor sn \rfloor}U_i, \quad s\in [0, 1]
\end{align}

where $U_i$ is a sequence of independent random variables, each taking values $1$ and $-1$ with probabilities $p\in [0, 1]$ and $1-p$ respectively. Within the interval, the input is a random walk with steps of size $w$

\begin{align}
    wX_{\lfloor sn \rfloor} =  w\sum_{i=1}^{\lfloor sn \rfloor} U_i
\end{align}

By the end of the interval the net input is simply the net difference of excitatory and inhibitory spikes. For instance, if the neuron receives $N$ excitatory and $M$ inhibitory spikes we get

\begin{align}
    v(t+\Delta t) &= \alpha v(t) + (1-\alpha)I + w (N-M)
\end{align}

 We are interested in computing the probability that the membrane potential crosses the threshold at some point within the timestep given that at the end of the timestep interval a total of $n=N+M$ spikes were received ($N$ excitatory and $M$ inhibitory) and the net integrated input was $k=N-M$. 

We will show that this is equivalent to asking what is the probability that the running maximum of a random walk is above a certain value $y$ given that the end value after $n$ steps is $k$. 

%


\subsubsection{Running maximum of a random walk}

With a slight abuse of notation we will refer to the random walk $X_{\lfloor sn \rfloor}$ as $X_n$

\begin{align}
    X_n =  \sum_{i=1}^{n} U_i, \quad n\in \mathbb{N}
    \label{eq:05_03_random_walk}
\end{align}

Let also $Y_n\!=\!\max \{X_1, X_2, \ldots, X_n\}$, the running maximum during the first $n$ steps. Finding the joint probability distribution $\mathbb{P}(Y_n\geq y, X_n=k)$ is a well known problem in the random walk literature which can be easily solved if we further assume that excitatory and inhibitory spikes arrive with the same probability $p\!=\!0.5$. In this case, the reflection principle can be used to obtain the following \cite{ibe2013markov, siegrist2021probability}

\begin{align}
    \mathbb{P}(Y_n\geq y, X_n=k) =  \mathbb{P}(X_n=2y-k), \quad k \leq y \leq n
    \label{eq:05_03_reflectionA}
\end{align}

The case where the running maximum is less than the final value ($y < k$) is easily obtained by noticing that the event $Y_n\geq y \subset X_n=k$. Leading to

\begin{align}
    \mathbb{P}(Y_n\geq y, X_n=k) =  \mathbb{P}(X_n=k), \quad y < k \leq n
    \label{eq:05_03_reflectionB}
\end{align}

As shown in \cite{siegrist2021probability}, these probabilities can be easily computed with

\begin{align}
    \mathbb{P}(X_n=k) &= 
    \begin{cases}
        \binom{n}{\frac{n+k}{2}} 2^{-n}, &\quad k\in \{\-n, -n+2, \ldots, n-2, n\} \\
        0, &\quad \text{otherwise}
    \end{cases}
    \label{eq:05_03_prob_final_value}
\end{align}

\subsubsection{Correcting the probability of firing with a random walk}

\begin{assumption} \label{as:05_03_symmetry}
    (Symmetric random walk) The probability of receiving a excitatory spike is the same as that of receiving an inhibitory spike.
\end{assumption}

With the previous results we can formulate the following theorem:

\begin{restatable}[]{thm}{05_03_random_walk_correct} \label{th:05_03_random_walk_correct}
Provided Assumptions \ref{as:05_03_weight}, \ref{as:05_03_deterministic} and \ref{as:05_03_symmetry} are satisfied. The probability that a LIF neuron with time constant $\tau$ fires at least once in a timestep of length $\Delta t$ given that it receives $N$ excitatory and $M$ inhibitory spikes each contributing $\pm w$ and an external current $I$ is given by

\begin{align}
    \mathbb{P}\left(\max_{s\in[t, t+\Delta t]}v(s)\geq V_{th} \: \Big| \:  v(t+\Delta t) \right) &= 
    \begin{cases}
        \frac{\mathbb{P}\left( X_n=2y - (N-M) \right)}{\mathbb{P}\left( X_n=N-M \right)}, &\quad N-M \leq y \leq n \\
        1, &\quad y < N-M \leq n \\
        0, &\quad y>N
    \end{cases}
    \label{eq:05_03_random_walk_correct}
\end{align}

where $n=N+M$, $y = \left\lceil \frac{V_{th}-(\alpha v(t)+(1-\alpha)I))}{w} \right\rceil$, with $\alpha=\exp(-\Delta t / \tau)$ and $X_n$ being a symmetric random walk.
\end{restatable}

\begin{proof*} 

\phantom{.}

Cases 1 and 2:

The deterministic part of the membrane potential during the interval $[t, t+\Delta t]$ is $\alpha v(t) + (1-\alpha)I$. Then, the number of excitatory inputs that a LIF neuron at potential $\bar{v}$ needs to cross the threshold is given by 

\begin{align}
    y = \left\lceil \frac{V_{th}-(\alpha v(t)+(1-\alpha)I))}{w} \right\rceil
\end{align}

Thus, the probability of crossing the threshold is the same as the probability of the running maximum being greater than $y$

\begin{align}
     \mathbb{P}\left(\max_{s\in[t, t+\Delta t]}v(s)\geq V_{th} \right) = \mathbb{P}\left(Y_n = \max_{i}X_i \geq y \right)
\end{align}

Since we know that the random walk ends at $X_n=N-M$ we have

\begin{align}
     \mathbb{P}\left(\max_{s\in[t, t+\Delta t]}v(s)\geq V_{th} \Big| \: v(t+\Delta t) \right) &= \mathbb{P}\left(Y_n\geq y \Big| \: X_n=N-M\right) \nonumber \\
     &= \frac{\mathbb{P}\left(Y_n \geq y , X_n=N-M\right)}{\mathbb{P}\left(X_n=N-M\right)} \nonumber \\
     &= \begin{cases}
         \frac{\mathbb{P}\left(X_n=2y-(N-M)\right)}{\mathbb{P}\left(X_n=N-M\right)} , &\quad N-M \leq y \leq n \nonumber \\
         \frac{\mathbb{P}\left(X_n=N-M\right)}{\mathbb{P}\left(X_n=N-M\right)}=1 , &\quad  y < N-M \leq n
     \end{cases} \nonumber \\
\end{align}

Case 3: 

If the number of steps required is less than the total number of excitatory inputs $N<y$ then the neuron will never cross the threshold in this timestep. Thus, the probability of spike is $0$. \hfill\qedsymbol{}
\end{proof*}

Note that a regular simulation where we consider all spikes taking place at the beginning is equivalent to a probability of spike according to Cases 2 and 3 only. The correction we add corresponds to Case 1, where a neuron that did not spike according to the final value $v(t+\Delta t)$ still could have spiked in between.

\begin{figure}[!ht]
\centering
\includegraphics[width=0.8\textwidth]{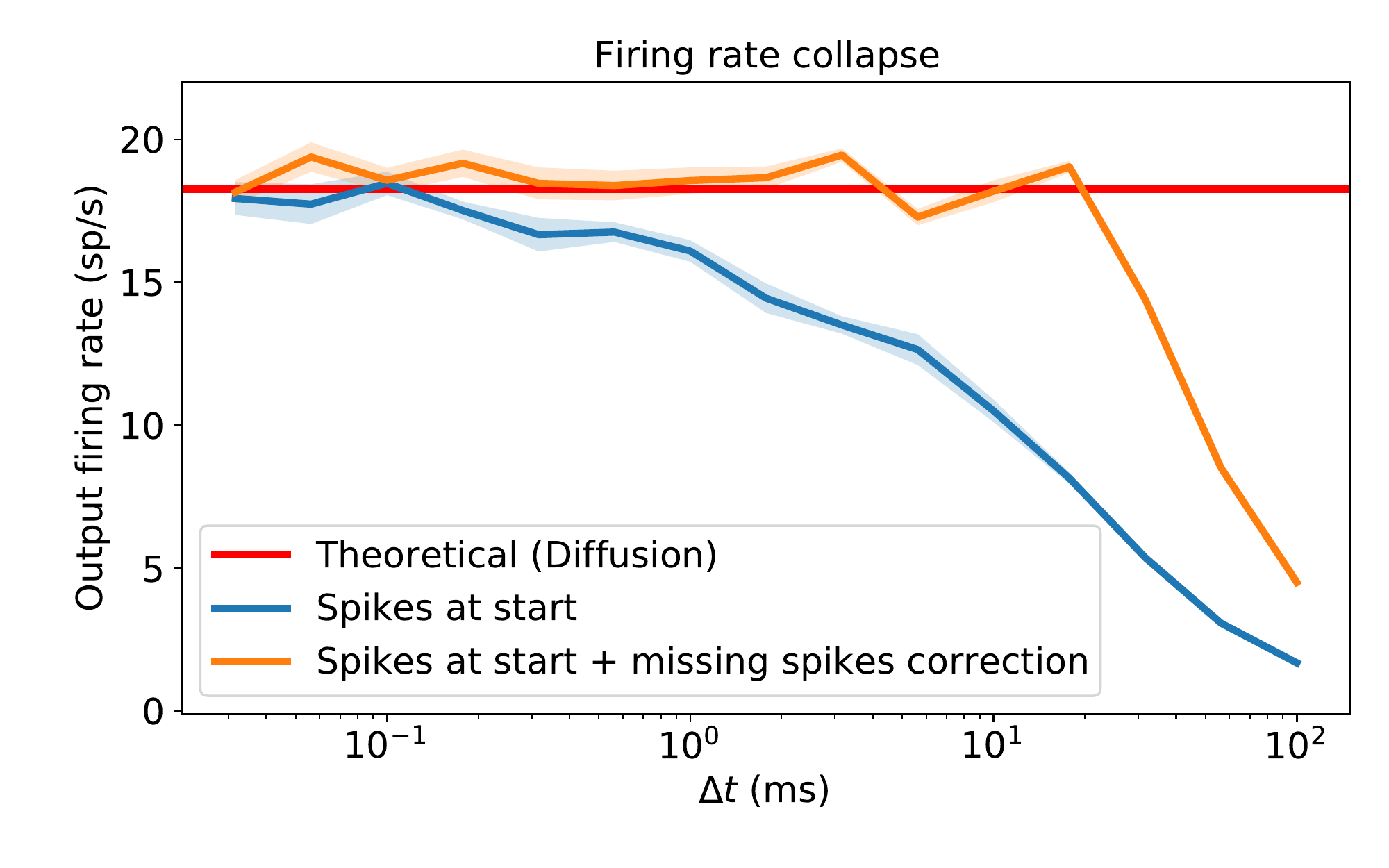}
\caption{Firing rate collapse as the simulation timestep increases and correction. Simulation details are identical to Figure \ref{fig:05_03_collapse}. The integration is performed identically but neurons that did not spike but could have spiked are made to spike randomly according to Theorem \ref{th:05_03_random_walk_correct}. Failure after $\Delta t\!=\!20ms$ is due to the $50$Hz Poisson generators no longer able to keep up.}
\label{fig:05_03_random_walk_correction}
\end{figure}

Implementing this correction is rather simple. We can perform the integration as usual and simply take those neurons that did not spike but could have spiked, compute their probability of spiking according to \eqref{eq:05_03_random_walk_correct} and randomly sample whether they spike. This is shown in Figure \ref{fig:05_03_random_walk_correction}. Failure after $\Delta t\!=\!20ms$ is due to the $50$Hz Poisson generators no longer able to keep up (as in Hypothesis \ref{hyp:05_03_hyp1}).

If we were to use the average value of $\alpha$ for the computation of $y$ we get the slightly worse result in Supplementary \ref{app:07_supp_ch5}. 

We can easily extend the result in Theorem \ref{th:05_03_random_walk_correct} to time varying thresholds $V_{th}(t)$ and non-linear membrane potential equations with the following result.

\begin{restatable}[]{thm}{05_03_random_walk_correct_nl}
Under the same assumptions as Theorem \ref{th:05_03_random_walk_correct}, a non-linear membrane equation of the form $\frac{dv}{dt}=f(v)+\sum_i w_i\delta(t-t_i)$ where the deterministic part is discretised with $g(v)$ and $w_i\in\{+w, -w\}$, and given a time-varying threshold $V_{th}(t)$ the results in Theorem \ref{th:05_03_random_walk_correct} hold for $y = \left\lceil \frac{V_{th}(t)-g(v(t))}{w} \right\rceil$.
\end{restatable}

\begin{proof*}
The proof is identical to that of Theorem \ref{th:05_03_random_walk_correct} upon substitution of the threshold and discretisation expressions. \hfill\qedsymbol{}
\end{proof*}

\subsection{Input spikes as a Wiener process}

An important limitation of the previous results is that it restricts us to only use two symmetric excitatory and inhibitory weights as well as requiring an identical excitatory and inhibitory rates. A simple workaround this can be achieved by considering that the number of incoming spikes during a timestep is large enough so that we can approximate the (now not necessarily symmetric) random walk with a Wiener process. This is similar to the diffusion approximation requirements with the subtle difference that the input rate must be high enough to ensure that number of spikes received in a timestep is large enough.

\subsubsection{Wiener Process as a limit of a random walk}

A Wiener process $W_t$ is a random process characterised by 

\begin{enumerate}
    \item $W_0 = 0$
    \item It has independent increments. $W_{t+s}\!-\!W_t, s\!\geq\! 0$ is independent from $W_u, u\!<\!t$
    \item It has Gaussian increments $W_{t+s}-W_t \sim \mathcal{N}(0, s)$
    \item It is continuous in $t$.
\end{enumerate}

The probability density function of a Wiener process follows a Gaussian distribution with mean $0$ and variance $t$

\begin{align}
     f_{W_t} (w) = \frac{1}{\sqrt{2\pi t}}\exp\left(-\frac{w^2}{2t} \right)
     \label{eq:05_03_wiener_pdf}
\end{align}

A standard result due to Donsker \cite{donsker1951invariance} states that given a sequence of independent identically distributed random variables $\xi_1, \xi_2, \ldots, \xi_n, $ with mean $0$ and variance $1$ the random walk:

\begin{align}
     W_n(t) = \frac{1}{\sqrt{n}} \sum_{1\leq k \leq \lfloor tn \rfloor} \xi_k, \quad t\in [0, 1]
\end{align}

converges in distribution to a Wiener process as $n\to \infty$. This result explains why Wiener processes are good models for many stochastic phenomena in nature. In fact, the diffusion approximation we introduced in section \ref{sec:05_02_diff_shot} uses this result to approximate the input spikes as Gaussian noise.

\subsubsection{Running maximum of a Wiener process}

We define the running maximum of a Wiener process as

\begin{align}
     M_t = \max_{0\leq s \leq t} W_s
\end{align}

The reflection principle can also be used in a Wiener process to obtain the joint density function (see \cite{lalley2001mathematical}\cite{lalley2013brownian} for a derivation)

\begin{align}
     f_{M_t, W_t} (m, w) = \frac{2(2m-w)}{t\sqrt{2\pi t}}\exp\left(-\frac{(2m-w)^2}{2t} \right), \quad m\geq0, w\leq m
     \label{eq:05_03_max_wiener_joint_pdf}
\end{align}

With \eqref{eq:05_03_wiener_pdf} and \eqref{eq:05_03_max_wiener_joint_pdf} we can compute the conditional density function 

\begin{align}
     f_{M_t | W_t} (m|w) = \frac{f_{M_t, W_t} (m, w)}{f_{W_t} (w)} = \frac{2(2m-w)}{t}\exp\left(-\frac{2m(m-w)}{t} \right), \quad m\geq0, w\leq m
\end{align}

Then, we can compute the conditional probability

\begin{align}
    \mathbb{P}\left(M_t \geq m | W_t \right) = \exp\left( -\frac{2m(m-W_t)}{t} \right), \quad m>\max(0, W_t)
\end{align}

\subsubsection{Correcting the probability of firing with a Wiener process}

With the previous results we can formulate the following Theorem.

\begin{restatable}[]{thm}{05_03_wiener_correct} \label{th:05_03wiener_correct}
Provided Assumptions \ref{as:05_03_weight} and \ref{as:05_03_deterministic} are satisfied. Then, the probability that a LIF neuron with time constant $\tau$ receiving an external current $I$ fires at least once in a timestep of length $\Delta t$ given that the weight distribution has mean $\mu_w$ and variance $\sigma_w$, and that by the end of the interval a net contribution of $W$ due to the incoming spikes was integrated, converges in distribution to 

\begin{align}
    \mathbb{P}\left(\max_{s\in[0, 1]}v(t+s\Delta t)\geq V_{th} \: \Big| \: v(t+\Delta t) \right) &= 
    \begin{cases}
        \exp\left( -\frac{2m(m-W_{t+\Delta t})}{\Delta t} \right), &\quad m>\max(0, W_{ t+\Delta t}) \\
        1, &\quad m \leq W_{t+\Delta t} \\
    \end{cases}
    \label{eq:05_03_wiener_correct}
\end{align}

as the number of input spikes per interval $n$ increases, with $m=\frac{V_{th}-(\alpha v(t)+(1-\alpha )I)-\mu_w))}{\sigma_w\sqrt{n/\Delta t}}$ and $W_{t+\Delta t}=\frac{W-\mu_w}{\sigma_w \sqrt{n/\Delta t}}$.
\end{restatable}

\begin{proof*} 

\phantom{.}

The membrane potential during the timestep is given by

\begin{align} \label{eq:05_03_lif_simplified_proof}
    v(t+s\Delta t) = \alpha v(t) + (1-\alpha)I +  \sum_{1\leq i\leq \lfloor sn \rfloor}w_i, \quad s\in [0, 1]
\end{align}

We can rewrite the stochastic part of the membrane potential as 

\begin{align}
    \sum_{1\leq i\leq \lfloor sn \rfloor}w_i = \mu_w +\sigma_w \sum_{1\leq i\leq \lfloor sn \rfloor}\xi_i
\end{align}

where now $\xi_i$ has zero mean and unit variance. Then, according to Donsker Theorem \cite{donsker1951invariance}, 

\begin{align}
    W_n(s) = \frac{1}{\sqrt{n}}\sum_{1\leq i \leq \lfloor sn \rfloor} \xi_i, \quad s\in [0,1]
\end{align}

converges in distribution to a Wiener process $W_{s}$ as the number of input spikes tends to infinity. Since our interval has a length of $\Delta t$ we simply have to rescale the process $W_{t+s\Delta t}=\sqrt{\Delta t} W_{s}$ and define $W_{t}=0$. Then, we can rewrite \eqref{eq:05_03_lif_simplified_proof} as 

\begin{align}
    v(t+s\Delta t) = \alpha v(t) + (1-\alpha)I + \mu_w +\sigma_w \sqrt{\frac{n}{\Delta  t}} W_{t+s\Delta t}, \quad s\in[0, 1]    
    \label{eq:05_03_proof_wienerB}
\end{align}

This means that for the neuron to spike we need the running maximum of the Wiener process of be larger than

\begin{align}
    m = \frac{V_{th}-(\alpha v(t) + (1-\alpha)I)-\mu_w }{\sigma_w \sqrt{n/\Delta t}}
\end{align}

That is

\begin{align}
    \mathbb{P}\left(\max_{s\in [0, 1]} v(t+s\Delta t) \geq V_{th} \right) &= \mathbb{P}\left(\max_{s\in [0, 1]} W_{t+s\Delta t} \geq  m \right)
\end{align}

Finally, since we know that at the end of the interval a net contribution of $W$ has been integrated we have that 

\begin{align}
    W_{t+\Delta t}=\sqrt{\Delta t} W_n(t+\Delta t) = \frac{W-\mu_w}{\sigma_w \sqrt{n/\Delta t}}
\end{align}

Which leads to 

\begin{align}
    \mathbb{P}\left(\max_{s\in [0, 1]} v(t+s\Delta t) \geq V_{th} \Big|  v(t+\Delta t)\right) &= \mathbb{P}\left(\max_{s\in [0, 1]} W_{t+s\Delta t} \geq  m \Big| W_{t+\Delta t}\right) \nonumber \\
    &= \begin{cases}
        \exp\left( -\frac{2m(m-W_{t+\Delta t})}{\Delta t} \right), &\quad m>\max(0, W_{t+\Delta t}) \\
        1, &\quad m \leq W_{t+\Delta t} \\
    \end{cases}
\end{align}

The second case is evident from the fact that if $m \leq W_{t+s\Delta t}$ then the final value is a running maximum and it is above the threshold. \hfill\qedsymbol{}
\end{proof*}

\begin{figure}[H]
\centering
\includegraphics[width=0.8\textwidth]{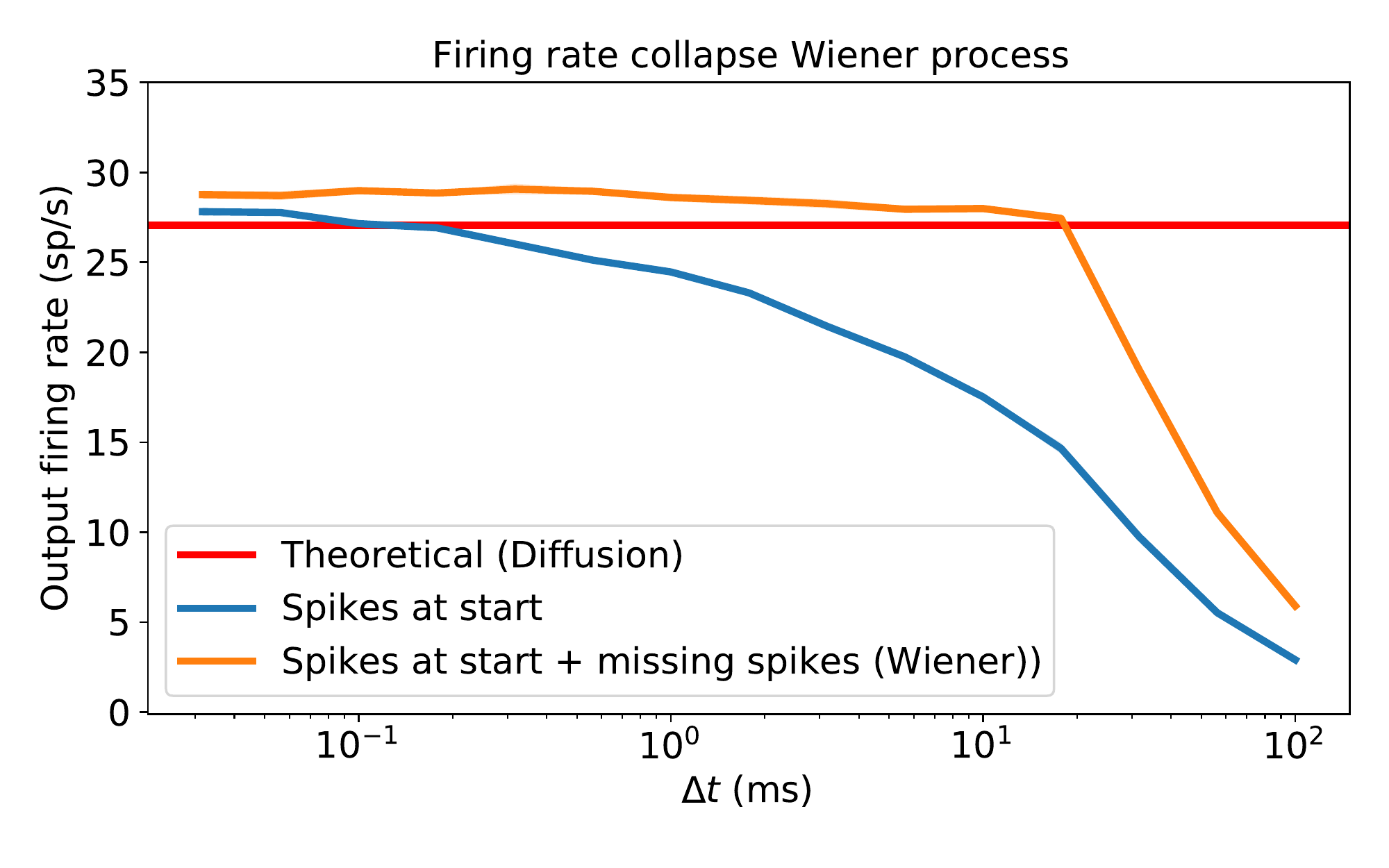}
\caption{Firing rate collapse as the simulation timestep increases and correction. Average and error in the mean for 5 samples are plotted. . Simulation details are identical to Figure \ref{fig:05_03_collapse} but now there are $2000$ inputs, the probability of connection is 1 and the weights are drawn from a Gaussian distribution with mean $0$ and standard deviation $0.01$. The integration is performed identically but neurons that did not spike but could have spiked are made to spike randomly according to Theorem \ref{th:05_03wiener_correct}. Failure after $\Delta t\!=\!20ms$ is due to the $50$Hz Poisson generators no longer able to keep up.}
\label{fig:05_03_wiener_correction}
\end{figure}

We can easily apply this correction by integrating as usual and taking the neurons that did not spike, compute their probability of spiking according to \eqref{eq:05_03_wiener_correct} and randomly sample whether they spike. This is shown in Figure \ref{fig:05_03_wiener_correction}. Failure after $\Delta t\!=\!20ms$ is due to the $50$Hz Poisson generators no longer able to keep up (as in Hypothesis \ref{hyp:05_03_hyp1}). We also observe that the correction is worse for small time steps as we would expect since there are not enough input spikes per timestep.

As in the random walk case, this result can be extended to more complex membrane models and time-varying threshold.

\begin{restatable}[]{thm}{05_03_wiener_correct_nl}
Under the same assumptions of Theorem \ref{th:05_03wiener_correct}, a non-linear membrane equation of the form $\frac{dv}{dt}=f(v)+\sum_i w_i\delta(t-t_i)$ where the deterministic part is discretised with $g(v)$ and given a time-varying threshold $V_{th}(t)$ the results in Theorem \ref{th:05_03wiener_correct} hold for $m = \frac{V_{th}(t)-g(v(t))-\mu_w}{\sigma_w \sqrt{n}}$.
\end{restatable}

\begin{proof*}
The proof is identical to that of Theorem \ref{th:05_03wiener_correct} upon substitution of the threshold and discretisation expressions. \hfill\qedsymbol{}
\end{proof*}

\subsection{General case}

For the general case where we do not make any assumption about either the weights or the input rates we cannot easily derive an analytical expression. Yet, we can design an algorithm where we sample a permutation of the input spikes.

\begin{algorithm}
\caption{A simulation of a single time step update for the general case}\label{alg:05_03_general_case}
\begin{algorithmic}
\State $v_{det} = \alpha v[t] + (1-\alpha)I$
\State $v[t+1] = v_{det} + \sum_{i=1}^nw_i$
\If {$v[t+1]\geq V_{th}$}
    \State The neuron spikes
    \State $v[t+1]=V_r$
\Else 
    \State $v_{sto}=$\texttt{accumulate}(\texttt{permute}($[w_1, w_2, \ldots, w_n]$))
    \If {\texttt{any}($v_{det}+v_{sto}\geq V_{th}$)}
        \State The neuron spikes
        \State $v[t+1]=V_r$
    \Else 
        \State The neuron does not spike
    \EndIf
\EndIf
\end{algorithmic}
\end{algorithm}

The function \texttt{accumulate} simply performs a cumulative sum of the input sequence. 

\begin{figure}[!ht]
\centering
\includegraphics[width=0.8\textwidth]{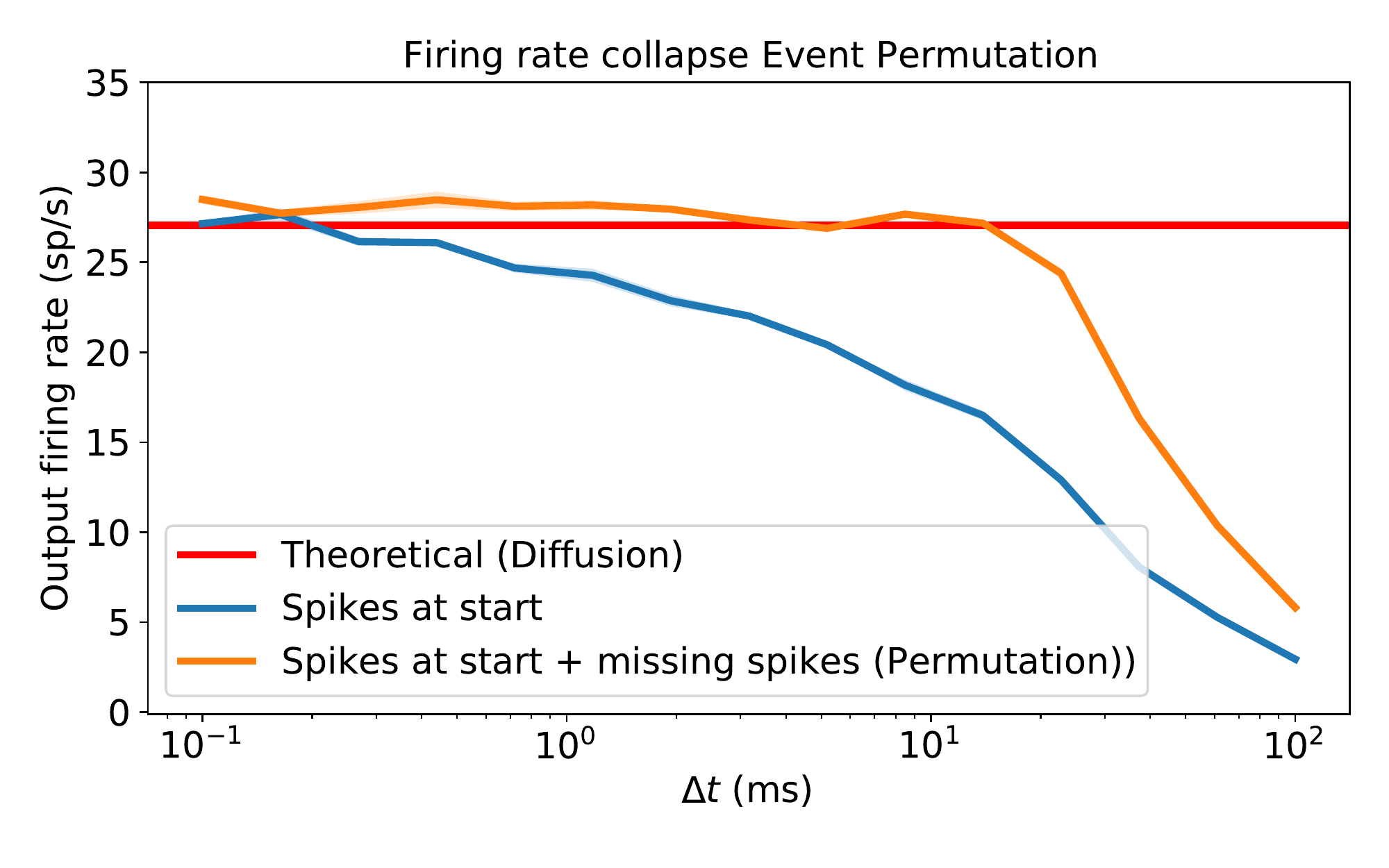}
\caption{Firing rate collapse as the simulation timestep increases and correction. Average and error in the mean for 5 samples are plotted. Simulation details are identical to Figure \ref{fig:05_03_wiener_correction}. The integration is performed according to Algorithm \ref{alg:05_03_general_case}. Failure after $\Delta t\!=\!20ms$ is due to the $50$Hz Poisson generators no longer able to keep up.}
\label{fig:05_03_permutation_correction}
\end{figure}

We can parallelise this algorithm to be executed for several neurons by simply zero padding the input weights to meet the sequence of the neurons which received the most inputs in that time step. The results are shown in Figure \ref{fig:05_03_permutation_correction}.

\section{Vanishing-Exploding gradients on SNNs} \label{sec:05_04_vanish_explode}

After correcting the rate collapse problem, the Fokker-Plank or shot-noise approaches should be enough to ensure a proper network initialisation in the forward pass. Yet, adjusting the network firing rates is not enough to ensure that the gradients will not vanish or explode. To this end, we now attempt to use the the variance propagation method for ANNs introduced in section \ref{subsec:05_02_var_prop} on SNNs.

Firstly, while strictly not necessary, we still include a section on applying this method to the forward pass and show that we can only get a weak bound on the firing rate at the output layer. Secondly, we attempt the same for the backward pass and obtain a closed form expression that depends on the membrane potential distribution which we can compute either by Fokker-Plank or threshold integration.

\subsection{Variance propagation in the forward pass of an SNN}

We consider a total of $L$ spiking layers with $n^{(l)}$ neurons in each layer $l=0, \ldots,  L\!-\!1$.  We use the unrolled membrane and spike expressions in discrete time

\begin{align}
v_j^{(l+1)}[t+1] &= \sum_{k=0}^{t} \alpha^{t-k} \left( \sum_i w^{(l)}_{ij}s^{(l)}_i[k] + (1-\alpha) RI \right) \\
s^{(l+1)}_j[t+1] &= H(v^{(l+1)}_j[t+1]-V_{th})
\end{align}

We adopt the following assumptions:

\begin{assumption}\label{as:05_04_fw1}
Membrane potentials $v_i^{(l)}$ across neurons at the same layer are independent and identically distributed.
\end{assumption}
\begin{assumption}\label{as:05_04_fw2}
The weight distribution is symmetric and $E[w_{ij}^{(l)}]=0,\: \forall i, j, l$ 
\end{assumption}
\begin{assumption}\label{as:05_04_fw3}
Weights $w_{ij}^{(l)}$ and spikes $s^{(l)}_i[k]$ at layer $l$ are independent.
\end{assumption}
\begin{assumption}\label{as:05_04_fw4}
Spike times are uncorrelated at different time steps, implying $Cov[s_i^{(l)}[k], s_i^{(l)}[p]]=Var[s_i^{(l)}[k]]\delta_{pk}$.
\end{assumption}
\begin{assumption}\label{as:05_04_fw5}
The probability of spike is the same and constant for neurons in the same layer and given by $P(s^{(l)}_i[k]=1)=\rho^{(l)}$ and $P(s^{(l)}_i[k]=0)=1-\rho^{(l)}$.
\end{assumption}

Assumptions \ref{as:05_04_fw1} through \ref{as:05_04_fw4} are analogous to the assumptions used in the variance propagation in ANNs as in \ref{subsec:05_02_var_prop} and \cite{he2015delving}.

Assumption \ref{as:05_04_fw5} implies that, given the timestep $\Delta t$ is small enough, all spike trains in the network are Poisson with a rate $r^{(l)}$ obeying $\rho^{(l)}\!=\!r^{(l)}\Delta t$. Note that this assumption would mean that the network cannot carry information in the precise spike times, however, this is only required for initialisation (i.e. before any training has taken place).

With these assumption we can formulate the following result:

\begin{restatable}[]{thm}{thFwd}
Given assumptions \ref{as:05_04_fw1}, \ref{as:05_04_fw2}, \ref{as:05_04_fw3} and \ref{as:05_04_fw4} are satisfied and further assuming that the membrane potential distribution is symmetric around zero, the probability that a neuron fires within a timestep is bounded according to 

\begin{align}
    \rho^{(l+1)}< \frac{n^{(l)}Var\left[ w^{(l)}_{ij} \right]}{2V_{th}^2} \left( A(\alpha)\rho^{(l)} + B(\alpha)\left(\rho^{(l)}\right)^2\right)
    \label{eq:05_05_forward_ineq}
\end{align}

with $A(\alpha) = \frac{1}{1-\alpha^2}$, $B(\alpha) = \frac{2\alpha}{(1-\alpha)^2(1+\alpha)}$.
\end{restatable}

\begin{proof*}
The proof can be found in Supplementary \ref{app:07_proofs_ch5}
\end{proof*}

There are several issues with this approach. Firstly, the symmetry of the membrane potential distribution assumption is in general too strong (see \ref{sec:05_02_diff_shot}) and will not be true for most parameter combinations. Secondly, this approach only offers an upper bound on the firing rate, not a lower bound, meaning we can only ensure the firing rate in the final layer will be lower than a certain value. This may of course lead to no firing at all. Finally, as we add more layers, the bound becomes weaker and weaker and thus it becomes less useful. 

It can be argued that other inequalities like Canteli's inequality do not require the symmetry assumption, however, it is weaker than Chebyshev's and thus the third issue remains. Even if we use other stronger inequalities we would still have the second issue, namely, we only bound from above. 


The problem stems from the fact that we are trying to find a relationship between the variance of the membrane potential and the firing rate. However, we know that this is not possible without considering the boundary conditions due to the resetting as we saw in \ref{sec:05_02_diff_shot}. 

In brief, the variance propagation method applied to the forward pass on SNNs does not give a useful solution to the network initialisation problem and other methods based on the Fokker-Plank equation or shot-noise are required instead.

\subsection{Variance propagation in the backward pass of an SNN}

We assume that the loss function applied to the last layer $\varepsilon : \mathbb{R}^{n^{(L-1)}} \to \mathbb{R}$ is a function of the sum of membrane potentials in the last layer across time (i.e.  $\varepsilon(v^{(L-1)}[t])\!=\!\mathcal{L}\left(\sum_t v^{(L-1)}[t]\right)$). For clarity, we will use $\frac{\partial \varepsilon}{\partial z}:= \Delta z$ in this section as in \cite{he2015delving}. The gradients of the loss function with respect the spikes at layer $l$ are

\begin{align}
    \Delta s^{(l)}[t] &= \sum_{k>t} \alpha^{k-t-1} \Delta v^{(l+1)}[k] W^{(l)}  \\
    \Delta v_i^{(l+1)}[t] &= \Delta s_i^{(l+1)}[t] H'\left(v_i^{(l+1)}[t]\right)
    \label{eq:05_04_SNN_gradient}
\end{align}

where $ \Delta s^{(l)}[t]\in\mathbb{R}^{1\times n^{(l)}}$, $\Delta v^{(l+1)}[t]\in\mathbb{R}^{1\times n^{(l+1)}}$, $W^{(l)}\in \mathbb{R}^{n^{(l+1)}\times n^{(l)}}$ and $H'(\cdot)$ is a function that subtracts the threshold $V_{th}$ and then applies the surrogate derivative of the Heaviside step function. See Supplementary \ref{app:07_grad_snn} for a derivation of these expressions.

We adopt the following assumptions:

\begin{assumption}\label{as:05_04_bw2}
The weight distribution has expectation $E[w_{ij}^{(l)}]=0,\: \forall i, j, l$ 
\end{assumption}
\begin{assumption}\label{as:05_04_bw1}
    The weights  $w_{ij}^{(l)}=(W^{(l)})_{ij}$ and the gradients $\Delta v^{(l+1)}[t]$ are independent.
\end{assumption}
\begin{assumption}\label{as:05_04_bw5}
     The gradient $\Delta s_i^{(l+1)}[t]$ and the surrogate $H'\left(v_i^{(l+1)}[t]\right)$ are independent.
\end{assumption}
\begin{assumption}\label{as:05_04_bw3}
    Gradients $\Delta s_i^{(l)}[t]$ that are far apart in time are independent.
\end{assumption}
\begin{assumption}\label{as:05_04_bw4}
    The surrogate function decreases fast enough so that if $H'(v_i^{(l+1)}[t])\!\neq\!0$ at a given $t$ it will be close to $0$ at nearby steps.
\end{assumption}

Assumptions \ref{as:05_04_bw1} and \ref{as:05_04_bw5} are of course incorrect since the weights at layer $l$ are used to compute the loss function and consequently gradients of the loss cannot be independent from the weights. Yet, these assumptions are commonly adopted in the weight initialisation literature with good empirical results \cite{he2015delving}.

Intuitively assumptions \ref{as:05_04_bw3} and \ref{as:05_04_bw4} mean that either neurons are not active (i.e. not close to the spiking threshold) for more than a single time step or in case they are the surrogate gradient is comparably very small at contiguous steps. These assumptions together with the fact that $\Delta v_i^{(l+1)}[k]=\Delta s_i^{(l+1)}[k]H'(v_i^{(l+1)}[k])$ mean that $Cov\left[\Delta v_i^{(l+1)}[k],\Delta v_i^{(l+1)}[p]\right]=Var\left[\Delta v_i^{(l+1)}[k]\right]\delta_{kp}$.

\begin{restatable}[]{thm}{thBwd}
Provided assumptions \ref{as:05_04_bw1} through \ref{as:05_04_bw5} are satisfied, the variance of the gradient of the loss function with respect to the spikes is given by 

\begin{align}
    Var\left[\Delta s_j^{(l)}[t]\right] = \frac{\rho_a^{(l+1)} }{1-\alpha^2} n^{(l+1)} Var\left[w_{ij}^{(l)}\right] Var\left[\Delta s_i^{(l+1)}[t]\right]
    \label{eq:05_04_backward_var}
\end{align}

where $\rho_a^{(l)}=E\left[\left(H'\left(v_i^{(l+1)}[t]\right)\right)^2\right]$.

\end{restatable}

\begin{proof*}
The proof can be found in Supplementary \ref{app:07_proofs_ch5}
\end{proof*}

This result is an expression of the variance of the gradients as a function of the variance of the next layer. With this, we can obtain an expression for the variance of the gradient on the first layer 

\begin{align}
    Var\left[\Delta s_j^{(0)}\right]
    &=  Var\left[\Delta s_j^{(L-1)}\right] \prod_{l=1}^{L-1} \frac{n^{(l)}Var\left[w_{ij}^{(l-1)}\right]\rho_a^{(l)}}{1-\alpha^2}
\end{align}

Which leads to setting 

\begin{align} \label{eq:05_04_optimal_var_bwd}
    Var\left[w_{ij}^{(l-1)}\right] &= \frac{1-\alpha^2 }{n^{(l)}\rho_a^{(l)}}
\end{align}

The challenge now is to find how to compute $\rho_a^{(l)}=E\left[\left(H'\left(v_i^{(l)}[t]\right)\right)^2\right]$.

\subsection{Computing \texorpdfstring{$\rho_a^{(l)}$}{ρ}}

We need to be careful when computing $E\left[\left(H'\left(v_i^{(l)}[t]\right)\right)^2\right]$. One may be tempted to think that we can obtain the probability density function of the membrane potential $P(v)$ in the steady-state using either the diffusion approximation or the threshold integration method and then simply integrate $\int_{-\infty}^{\infty} \left(H'\left(v\right)\right)^2 P(v) dv$

However, so far we have been using the membrane potential, $v$, in the diffusion, shot-noise and threshold integration methods as if it was identical to that defined for the neuron update (e.g. \eqref{eq:05_03_lif_correct}). These are not the same, however, since the membrane in neuron updates equation has not yet applied the threshold-and-reset mechanism. When we apply the threshold function to the membrane potential we do so \emph{after} updating it with the incoming spikes and currents within a timestep and right \emph{before} applying the reset. Thus, we do not apply the surrogate function to the steady state distribution but to what we will call the \textit{before-reset distribution}. That is, the distribution we obtain after a time step $\Delta t$ has taken place with initial condition $P(v)$ but before resetting the neurons.

For instance, in the diffusion case, in the absence of a spike-and-reset mechanism the membrane potential distribution evolves according to the Fokker Planck equation

\begin{align}
    \tau \frac{\partial P(v, t)}{\partial t}= &-\frac{\partial}{\partial v}\left(\left[ -v + \mu\right] P(v, t)\right)
    + \frac{\sigma^2}{2}  \frac{\partial^2}{\partial v^2}P(v, t)    
    \label{eq:05_04_diffussion_eq_no_reset}
\end{align}

To obtain the \textit{before-reset distribution} distribution we just need to find $\hat{P}(v)\!=\!P(v, t_0 + \Delta t)$ given an initial condition $P(v, t_0)=P(v)$. A similar result can be obtained using the shot-noise equations but instead of a partial differential equation (PDE) we would had a system of differential algebraic equations (DAE) which most numerical solvers struggle with, an exception being MATLAB's PDE solver\footnote{See \url{https://uk.mathworks.com/help/matlab/math/partial-differential-equations.html}}.

Once the before-reset distribution $\hat{P}(v)$ has been found we can simply compute 

\begin{align} \label{eq:05_05_general_rho}
    \rho_a^{(l)} = \int_{-\infty}^{\infty} \left(H'\left(v\right)\right)^2 \hat{P}(v) dv
\end{align}

For the special case where our surrogate function is given by $H'(v) = H(v-(V_{th}-B_{th}))-H(v-(V_{th}+B_{th}))$ as in \cite{perez2021sparse} we have

\begin{align}
    \rho_a^{(l)} &= \int_{V_{th}-B_{th}}^{V_{th}+B_{th}} \hat{P}(v) dv
\end{align}

Provided that $B_{th}$ is small as required by Assumption \ref{as:05_04_bw4}, we can interpret $\rho_a^{(l)}$ as the probability of the backward spike gradient to be non-zero or if we prefer that the backward gradient `fires' as in \cite{perez2021sparse}.

\begin{figure}[!ht]
\centering
\includegraphics[width=0.99\textwidth]{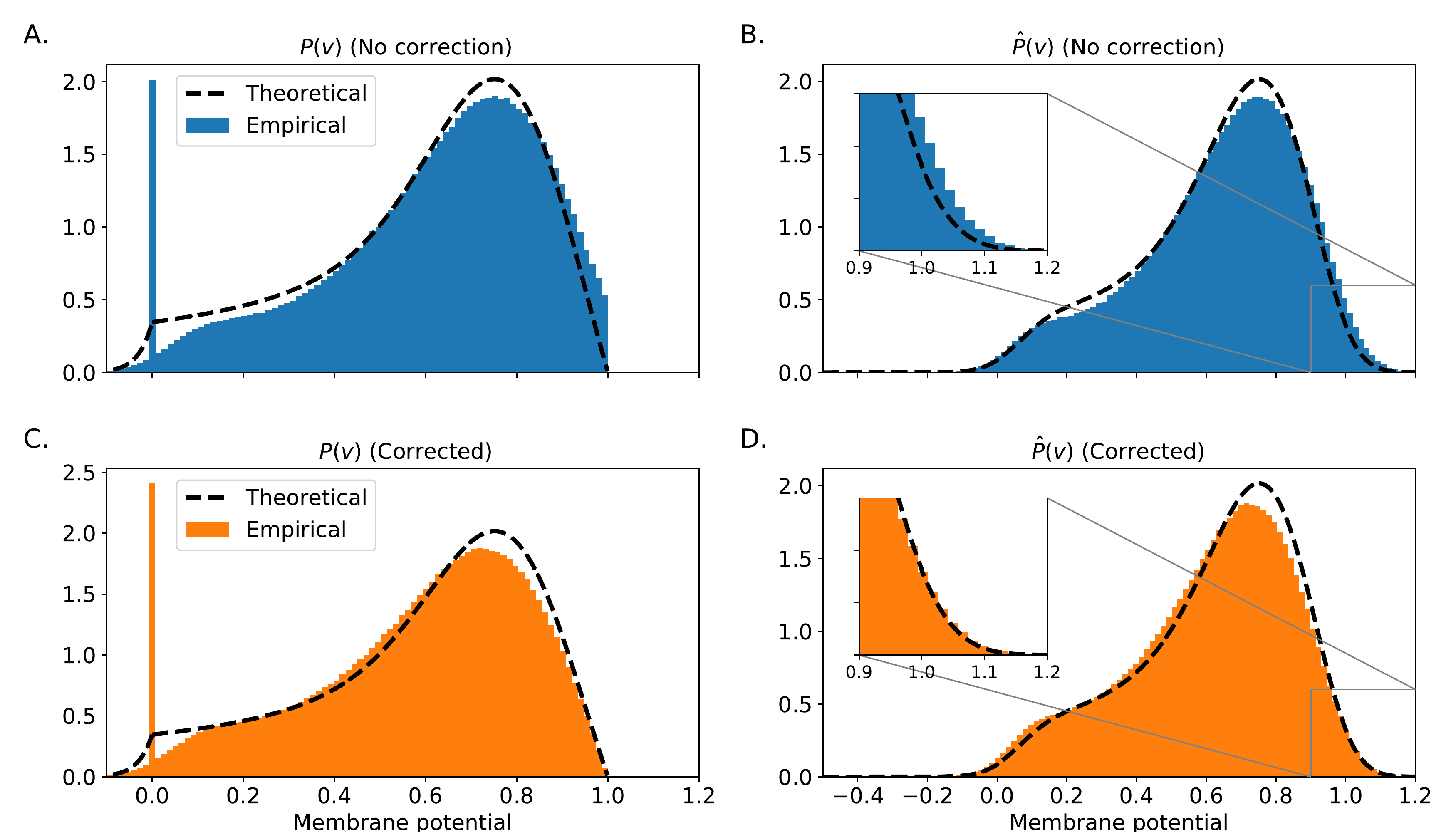}
\caption{Membrane distributions theoretical and empirical. The theoretical distribution $P(v)$ was obtained by using the membrane distribution equation according to the diffusion approximation \eqref{suppeq:05_02_diff_dist} and the  before reset distribution was obtained by solving \eqref{eq:05_04_diffussion_eq_no_reset} using the \texttt{py-pde} package \cite{zwicker2020pypde}. Note how the distribution on the corrected case fits much better around the threshold, leading to a better approximation of $\mathcal{I}$. The parameters for this simulations were $\Delta t\!=\!1$ms, $\tau\!=\!10$ms, $I_{ext}\!=\!0.9$, $R\!=\!1$, $V_r\!=\!0$, $V_{th}\!=\!1$, $\mu_{w}\!=\!0$, $\sigma_{w}\!=\!0.0096$ and $\nu_{in}\!=\!\nu_{target}=\!30$Hz.  }
\label{fig:05_04_distributions}
\end{figure}

Figure \ref{fig:05_04_distributions} shows the membrane potential distributions after reset ($P(v)$) and before reset ($\hat{P}(v)$) for both the Wiener corrected forward pass (bottom row) and the original forward pass (top row). Note how the corrected version results in a much better approximated distribution around the threshold, necessary for computing $\mathcal{I}$.

\section{General SNN Initialisation strategy}

The aim of this paper is to find a way to initialise the weights of an SNN to ensure a certain firing rate at the output while avoiding having exploding or vanishing gradients.
After the results from the previous sections we have the following:

\begin{enumerate}
    \item We can compute the firing rate at each layer by using either the diffusion or the shot noise approximation plus using an appropriate firing rate collapse solution if necessary.
    \item We have an expression for what the desired weight variance should be in order to avoid vanishing or exploding gradients.
\end{enumerate}

The question now is how can we use these results to obtain a desired firing rate while ensuring the gradients will flow.

Let $\mathcal{V}$ be a operator that computes the firing rate at a layer $l$ given the variance of the weights $\sigma^2_{w^{(l)}}$ and other parameters $\pmb{\theta}$ (threshold, input rates, time constant etc). This algorithm can be Siegert's integral, shot-noise rate equation or threshold integration.

\begin{align}
    \nu^{(l)} = \mathcal{V}(\sigma_w^2, \pmb{\theta})
\end{align}

Let $\mathcal{W}$ be the operator that computes the optimal weight variance $\hat{\sigma}_w^2$ for the gradients to flow. This operator will include several steps: computing the stationary distribution, computing the before-reset distribution and finally computing the weight variance with \eqref{eq:05_04_optimal_var_bwd}. Note that one of the input parameters of $\mathcal{W}$ will be the initial variance $\sigma^2_{w^{(l)}}$.

\begin{align}
    \hat{\sigma}_w^2 &= \mathcal{W}(\sigma_w^2, \pmb{\theta})
\end{align}

This two variances, $\sigma_w^2$ and $\hat{\sigma}_w^2$, will not necessarily be the same, but we would like for them to be equal. This is easily achieved by allowing the surrogate function to be scaled by a factor $\kappa H'(v)$. Then, equation \eqref{eq:05_05_general_rho} becomes

\begin{align} \label{eq:05_05_general_rho2}
    \rho_a^{(l)} &= \kappa^2 \int_{-\infty}^{\infty} \left(H'\left(v\right)\right)^2 \hat{P}(v) dv \nonumber \\
    &= \kappa^2 \mathcal{I}
\end{align}

substituting into \eqref{eq:05_04_optimal_var_bwd} the expression for the backward optimal variance becomes

\begin{align} \label{eq:05_04_optimal_var_bwd2}
    \sigma_w^2 &= \frac{1-\alpha^2 }{n^{(l)}\kappa^2 \mathcal{I}}
\end{align}

and finally we can obtain an expression for $\kappa$

\begin{align} \label{eq:05_04_optimal_kappa}
    \kappa &= \sqrt{\frac{1-\alpha^2 }{n^{(l)}\sigma_w^2 \mathcal{I}}}
\end{align}

We note that the integral $\mathcal{I}$ needs to be only computed once since it is fully determined by the forward pass.

Thus, the general strategy for initialising and SNN consists of the following two steps:

\begin{enumerate}
    \item Set a desired firing rate and obtain $\sigma_w^2$ via a gradient-free root finding algorithm to solve $\nu^{(l)} - \mathcal{V}(x, \pmb{\theta})=0$ for $x$. 
    \item Compute $\kappa$ according to equation \eqref{eq:05_04_optimal_kappa}.
\end{enumerate}

Since the surrogate gradient does not affect the computation of the forward firing rate it is safe to modify it without affecting the forward pass. As opposed to \cite{rossbroich2022fluctuation}, the scaling we apply to the surrogate function considers the statistics of the membrane potential (through $\mathcal{I}$), thus, we do not expect the gradients to explode as in their case. Moreover, since it has been found that surrogate gradient descent is remarkably robust to the surrogate function we expect that training will take place without issue \cite{zenke2021remarkable}.

\begin{figure}[!ht]
\centering
\includegraphics[width=0.49\textwidth]{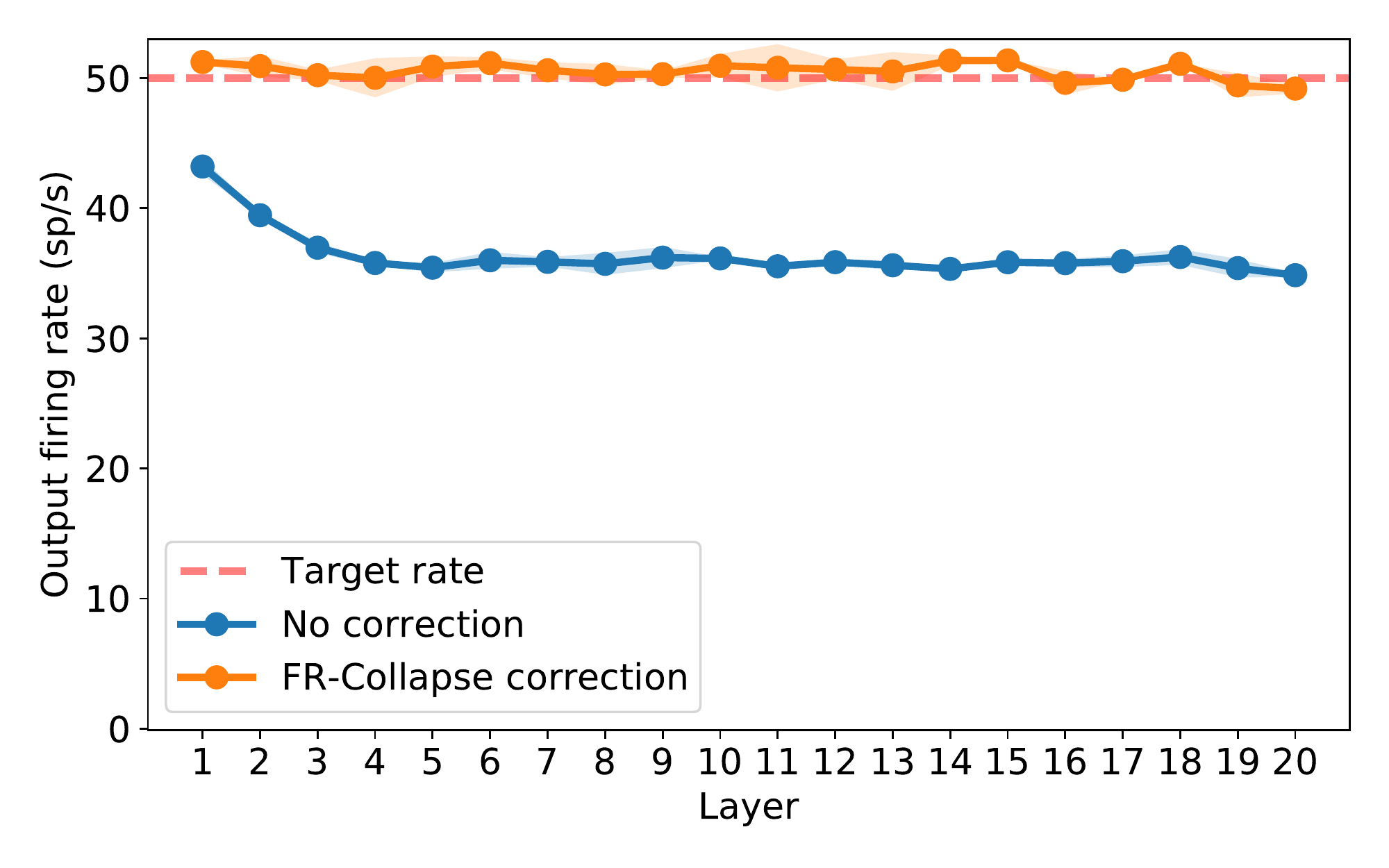}
\includegraphics[width=0.49\textwidth]{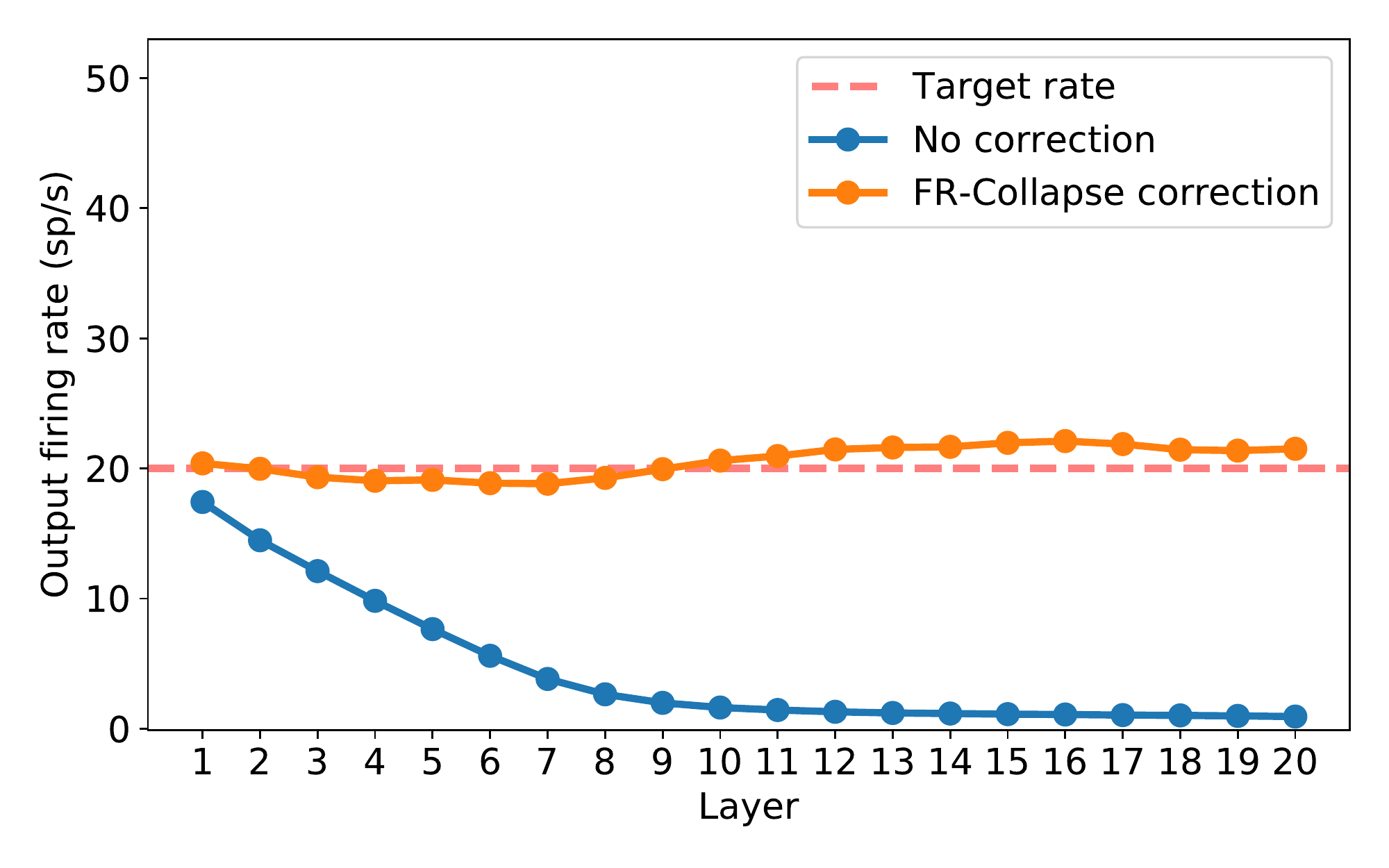}
\caption{Firing rate at each layer on a feed-forward SNN with 20 layers with target rates $50$Hz and $20$Hz respectively. We plot the mean and standard error in the mean for $5$ trials. The input spikes where Poisson trains at the target rate. Each layer consisted of $2000$ LIF neurons with parameters $\tau\!=\!10$ms, $I=\!0.6$, $V_r\!=\!0$, $V_{th}\!=\!1$. The probability of connection between neurons in consecutive layers was $0.5$. The operator $\mathcal{V}$ was the diffusion rate. The optimal weights found were $0.0387$ and $0.0299$ respectively. Optimisation was done using \texttt{scipy.root\_scalar} function with \texttt{toms748} method. The network was simulated for $1$s with timestep $\Delta t \!=\!1$ms.}
\label{fig:05_05_forward}
\end{figure}

We apply this method and obtain the results depicted in Figure \ref{fig:05_05_forward}. As we see, as long as the firing rate collapse correction is added (in this case the random walk version) the firing rate remains around the target rate quite reliably. Interestingly, at high enough rates the firing rate collapse does not go to zero but converges to some value (Figure \ref{fig:05_05_forward} left). It may be interesting to study this further as predicting this value could remove the need to use the firing rate collapse correction altogether. In addition, we found that trying lower target rates results in the network biasing towards a slightly higher rate (see Supplementary \ref{app:07_init_fail}) this also needs to be studied further. 

\begin{figure}[!ht]
\centering
\includegraphics[width=0.8\textwidth]{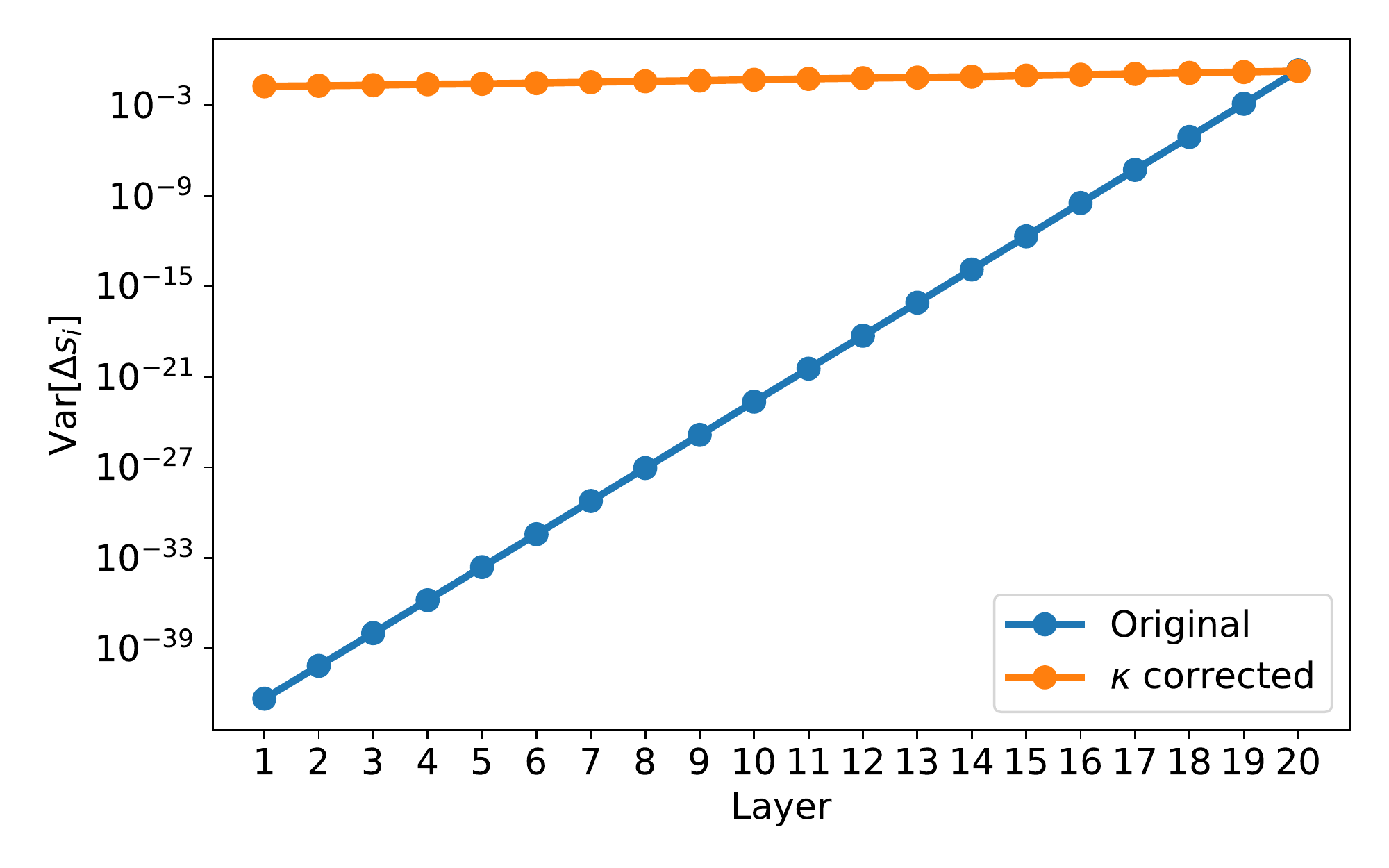}
\caption{Variance of the spikes gradient at each layer on a feed-forward SNN with 20 layers with target rate of $30$Hz. The input spikes where Poisson trains at the target rate. Each layer consisted of $2000$ LIF neurons with parameters $\tau\!=\!10$ms, $I=\!0.9$, $V_r\!=\!0$, $V_{th}\!=\!1$. The optimal weights found had standard deviation $\sigma_w\!=\!0.0096$. Optimisation was done using \texttt{scipy.root\_scalar} function with \texttt{toms748} method. The surrogate function used was piecewise-constant as in \cite{perez2021sparse} with $B_{th}=0.01$, and $g=\kappa$ (corrected) or $g=1$ (original). The network was simulated for $1$s with timestep $\Delta t \!=\!1$ms.}
\label{fig:05_05_backward}
\end{figure}

Finally, Figure \ref{fig:05_05_backward} shows the gradient at each layer in a network where the forward correction was done using the Wiener method. We compare the variance of the spike gradient $\Delta s_i$ with and without the $\kappa$ correction. As we see, the original gradient vanishes while the $\kappa$-corrected gradient remains nearly constant. We note however, that the variance still decays to some extent due to the fact that our assumptions are not completely correct. In fact, using a larger value of $B_{th}$ results in a much worse variance propagation due to Assumption \ref{as:05_04_bw4} not being true and consequently the covariance term between gradient being too large (see Supplementary \ref{app:07_init_fail}). Yet, as shown in \cite{perez2021sparse}, a small value of $B_{th}$ (e.g. $B_{th}\!=\!0.01$ as in the figure) is still perfectly fit for training successfully. 

\section{Discussion and Conclusions}

The development of new techniques that allowed to train deeper and deeper ANNs has been an integral component of the success of ANNs in recent years. Proper weight initialisation has been shown to be crucial to achieve the best accuracy and fast convergence on deep neural networks, especially when composed of many layers \cite{skorski2021revisiting}. Empirical evidence has shown that compared to ANNs, deep SNNs are even more sensitive to suboptimal weight initialisation where networks with just a few layers already pose a challenge \cite{rossbroich2022fluctuation}. 

Although there is a wide range of weight initialisation methods for ANNs, relatively simple techniques based on variance propagation are the norm in most works. However, we have shown that this method is insufficient for the forward propagation of SNNs for two reasons. Firstly, the relationship between the variance of the membrane potential and the variance of the spikes (or equivalently the probability of spiking) is not straight forward. Secondly, we observed that the probability of spiking seems to decrease as we decrease the resolution of the SNN simulation (the firing rate collapse problem).

The first problem has already been tackled in the past by developing techniques based in finding the stationary membrane distribution. These include the diffusion approximation \cite{gerstner2014neuronal}, shot-noise \cite{richardson2010firing} or threshold integration \cite{richardson2008spike}. These methods should be enough to allow to predict the output firing rate under a wide range of circumstances and neuron models. However, all of them are developed for ideal continuous-time simulations and do not consider the firing rate collapse problem.

The firing rate collapse problem has often been suggested to be just an inability of the neurons to fire enough times at low temporal resolutions. We have shown that this is not enough to fully explain this phenomenon, instead, the issue often arises from the fact that all inhibitory-excitatory inputs are cancelled instantaneously within a timestep. Thus, removing all output spikes that could take place at some permutation of the input spikes. We have shown that we can take this into account using different methods: random walk, Wiener noise or sampling a random permutation. 

While these methods solve the firing rate collapse problem, it is yet to be seen what impact they can have in SNN training. Since the newly added spikes are stochastic, this could lead to requiring more training steps to properly sample the network response. Yet, given that this is only important during initialisation, we could decrease the probability of firing as the training progresses and substitute this by a homeostatic rule as in \cite{zenke2021remarkable, perez2021sparse, rossbroich2022fluctuation}. Moreover, we also need to develop a method to evaluate how many layers can we stack while somewhat accurately predicting the final rate when using real data instead of simple homogeneous Poisson spikes. 

In addition, there is also the added computational cost which, in principle, should be small in the random walk and Wiener cases as the operations required are linear with the number of neurons ($O(n)$) and thus negligible compared to the synaptic cost ($O(n^2)$). For the permutation method, if properly implemented, the cost would be linear with the number of events. Since the number of events per timestep is bounded by the number of weights, this method would scale quadratically with the number of neurons ($O(n^2)$) and thus not increasing the overall computational complexity. 

We also obtained an expression for the variance propagation of the gradients in the backward pass and empirically tested its effectiveness. Our results suggest that it is possible to ensure for the gradients to remain bounded by scaling the surrogate function provided some assumptions are met. While most of the assumptions are equivalent to those made on the ANN literature, the last assumption \ref{as:05_04_bw4} is the most problematic one since it implies that backward surrogate gradients are uncorrelated. For this to be true we need to constrain the surrogate function to be very narrow. Previous work \cite{perez2021sparse} showed that it is indeed possible to use a very narrow gradient while still learning efficiently which gives reason to believe that it is reasonable to use this assumption. Besides, our empirical results suggest that the gradients do indeed remain nearly constant when using this correction. However, empirical evidence is needed to ensure that this actually works with real data. 

In this paper we have assumed that the inputs are homogeneous Poisson spike trains. In real datasets this is rarely the case and in particular most interesting datasets are neither homogeneous nor Poisson. This is a very important issue that needs to be tackled for these methods to be successful. Yet, our results as well as previous work on SNN initialisation on real datasets are encouraging \cite{rossbroich2022fluctuation}.

In brief, we have devised strategies to optimally initialise SNNs by considering their true membrane distribution and the problems arising from discrete time simulations of continuous systems. However, there is still the need to empirically prove their viability on real data.

\newpage
\beginsupplement
\section{Extended Background Theory} \label{supp_sec:05_02_background}

In this section we add more details and present more extended derivations on the background theory for readers that are unfamiliar with either ANN variance propagation for weight initialisation or diffusion, shot-noise and threshold integration methods. We begin deriving forward and backward variance propagation on ANNs for optimal weight initialisation and different methods to compute the output firing rate and membrane distribution of a layer of LIF neurons including: diffusion approximation, shot-noise and threshold integration.

\subsection{Initialisation based on variance flow in ANNs} \label{supp_subsec:05_02_var_prop}

The derivations in this section are a more detailed version of those in \cite{he2015delving}. We first compute the variance of the activations and then the variance of the gradients on feedforward/convolutional ANNs. From this we derive the optimal weight variance to keep the activations variance bounded. We explore the use of this method on SNNs in section \ref{sec:05_04_vanish_explode}.

We often use the following identities for random variables $X_i$ and scalars $a_i \in \mathbb{R}$:

\begin{align}
   \qquad Var\left[ \sum_{i=1}^N a_i X_i\right] = \sum_{i=1}^N a_i^2 Var\left[X_i\right] + 2 \sum_{1\leq i < j \leq N} a_ia_j Cov\left[X_i, X_j\right]
\end{align}

for the special case where $ X_i, X_j$ are uncorrelated $\forall \: i, j$ with $i\neq j$

\begin{align} 
    \qquad Var\left[ \sum_{i=1}^N a_i X_i\right] = \sum_{i=1}^N a_i^2 Var\left[X_i\right]
\end{align}

For independent random variables variables $X, Y$

\begin{align}    
    \qquad Var\left[XY\right] = E[X^2]E[Y^2]- \left(E[X]\right)^2 \left(E[Y]\right)^2
\end{align}

\subsubsection{Forward variance propagation on ANNs} \label{supp_sec:05_02_ann_fwd}
Unlike SNNs, ANNs do not have any dynamics or resetting present, they simply consist of an affine transformation followed by a non-linearity.

\begin{align}
    y^{(l+1)} &= W^{(l)}x^{(l)} + b^{(l)} \\
    x^{(l+1)} &= f(y^{(l+1)})
\end{align}

where $y^{(l+1)}, x^{(l+1)} \in \mathbb{R}^{n^{(l+1)}}$, $W \in \mathbb{R}^{{n^{(l+1)}}\times {n^{(l)}}}$, $x^{(l)} \in \mathbb{R}^{n^{(l)}}$, $b^{(l)}=\pmb{0}$ at initialisation, $f$ is a non-linear element-wise operation and $l=0, \ldots, L\!-\!1$. We make the assumptions that $x_i^{(l)}$ are i.i.d. for all $i$ and $l$, $W^{(l)}_{ij}$ and $x^{(l)}_{i}$ are independent and $W^{(l)}_{ij}$ is zero mean and symmetric around zero and i.i.d for all $i$, $j$ and $l$. Then

\begin{align}
    Var\left[y_j^{(l+1)} \right]&= Var\left[ \sum_i
    W^{(l)}_{ij}x^{(l)}_i+b_j^{(l)} \right] \nonumber \\ 
    &= n^{(l)}Var\left[  W^{(l)}_{ij}x^{(l)}_i \right] \nonumber \\ 
    &= n^{(l)}Var\left[  W^{(l)}_{ij} \right]E\left[ \left(x^{(l)}_i\right)^2 \right] 
\end{align}

Given that $W^{(l)}_{ij}$ is zero mean and symmetric so is $y^{(l+1)}_{i}$. Thus, for a typical case of a ReLU function $f(\cdot)=max(0, \cdot)$  we have

\begin{align}
    E\left[ \left(x^{(l)}_i\right)^2 \right] &= E\left[ \left( max(0, y_i^{(l)}) \right)^2 \right] \nonumber \\
    &= \frac{1}{2} E\left[ \left( y_i^{(l)} \right)^2 \right] \nonumber \\
    &= \frac{1}{2} Var\left[  y_i^{(l)}  \right]
\end{align}

Resulting in

\begin{align} \label{supp_eq:05_02_var_anns}
    Var\left[y_j^{(l+1)} \right]&= \frac{1}{2} n^{(l)}Var\left[  W^{(l)}_{ij} \right] Var\left[  y_i^{(l)}  \right]
\end{align}

and for $L$ layers

\begin{align}
    Var\left[y_j^{(L)} \right]&= Var\left[  y_i^{(1)} \right] \prod_{l=1}^{L-1} \frac{1}{2} n^{(l)}Var\left[  W^{(l)}_{ij} \right]
\end{align}

Ensuring that the variance of the activations does not increase or decrease exponentially with the number of layers, results in Kaiming's initialisation \cite{he2015delving}

\begin{align} \label{supp_eq:05_02_ann_init_forward}
    Var\left[ W^{(l)}_{ij} \right] = 2/n^{(l)}
\end{align}

\subsubsection{Backward variance propagation in ANNs} \label{supp_sec:05_02_ann_bwd}

Ideally gradients should propagate from the last layer to the first without exploding or vanishing. To achieve this we follow the same approach as before but now with the gradients. 

We begin by rewriting the forward pass equations

\begin{align}
    y^{(l+1)} &= W^{(l)}x^{(l)} + b^{(l)} \\
    x^{(l+1)} &= f(y^{(l+1)})
    \label{supp_eq:05_02_f(y)}
\end{align}

Where $y^{(l+1)}, x^{(l+1)} \in \mathbb{R}^{n^{(l+1)}}$, $W^{(l)} \in \mathbb{R}^{{n^{(l+1)}}\times {n^{(l)}}}$, $x^{(l)} \in \mathbb{R}^{n^{(l)}}$, $b^{(l)}=\pmb{0}$ at initialisation and $f$ is an non-linear element-wise operation and $l=0, \ldots, L\!-\!1$.

Given a loss function $\varepsilon: \mathbb{R}^{n^{(L\!-\!1)}} \to \mathbb{R}$ we can write the gradient of the loss with respect to the activations

\begin{align}
    \frac{\partial \varepsilon}{\partial x^{(l)}} &= \frac{\partial \varepsilon}{\partial y^{(l+1)}} \frac{\partial y^{(l+1)}}{\partial x^{(l)}}
\end{align}

where $\frac{\partial \varepsilon}{\partial x^{(l)}}\in\mathbb{R}^{1\times n^{(l)}}$, $\frac{\partial \varepsilon}{\partial y^{(l+1)}}\in\mathbb{R}^{1\times n^{(l+1)}}$ and $\frac{\partial y^{(l+1)}}{\partial x^{(l)}}\in\mathbb{R}^{n^{(l+1)}\times n^{(l)}}$ (note that we are using numerator layout convention). Using \eqref{supp_eq:05_02_f(y)} we can expand

\begin{align}
    \left( \frac{\partial \varepsilon}{\partial y^{(l+1)}} \right)_i &= \left( \frac{\partial \varepsilon}{\partial x^{(l+1)}} \right)_i f'\left(y_i^{(l+1)}\right)
\end{align}

We also note that $\frac{\partial y^{(l+1)}}{\partial x^{(l)}}=\frac{\partial  W^{(l)}x^{(l)}}{\partial x^{(l)}}=W^{(l)}$\footnote{While we are presenting the case of the fully connected layer, the following results also hold for other architectures such as convolutional networks by simply adjusting the backward weight matrix $\frac{\partial y^{(l+1)}}{\partial x^{(l)}}=\widehat{W}^{(l)}$. For instance in a regular convolutional layer in the forward pass, we have a filter of size $k\times k$ and a total of $c$ input channels and $d$ output channels: $x^{(l)}\in \mathbb{R}^{ck^2\times 1}$ and $y^{(l+1)}\in \mathbb{R}^{d\times 1}$ and $W^{(l)}\in \mathbb{R}^{d\times ck^2}$. I.e., $x^{(l)}$ is now an input block (a window spanning all input depths) which has been flattened, $y^{(l+1)}$ is an output block (a single neuron per output depth) and $W^{(l)}$ is the $d$ filters flattened and stacked. This operation is repeated for each input block. Thus, a neuron in the input affects a total of $dk^2$ neurons in the output (it helps to think of a single input arriving at a given output neuron by a moving window $k$ times in both height and width). This leads to  us defining a new $\hat{x}^{(l)}\in\mathbb{R}^{c\times 1}$, $\hat{y}^{(l+1)}\in\mathbb{R}^{dk^2\times 1}$ and $\widehat{W}^{(l)}\in\mathbb{R}^{k^2d\times c}$ and $W^{(l)}$ can be reshaped into $\widehat{W}^{(l)}$. Then the only difference in the results of this subsection are obtained by using $\hat{n}^{(l)}=dk^2$ instead of $n^{(l)}=ck^2$.}.

For notational simplicity, throughout this subsection we use $\frac{\partial \varepsilon}{\partial z }\!:=\!\Delta z$ for any partial derivative of the loss with respect to $z$ as in \cite{he2015delving}.

This means we end up with the following equations

\begin{align}
    \Delta x^{(l)} &= \Delta y^{(l+1)} W^{(l)} \\
    \Delta y_i^{(l+1)} &= \Delta x_i^{(l+1)} f'\left(y_i^{(l+1)}\right)
\end{align}

We make the assumptions that $\Delta y_i^{(l+1)}$ are i.i.d. for all $i$ and $l$, $W^{(l)}_{ij}$ and $\Delta y^{(l+1)}_{i}$ are independent and $W^{(l)}_{ij}$ is zero mean and symmetric around zero and i.i.d for all $i$, $j$ and $l$. We further assume that $\Delta x_i^{(l+1)}$ and $f' \left(y_i^{(l+1)}\right)$ are independent.

We want to compute the variance of the gradients of a neuron on a given layer $\Delta x^{(l)}_j$ as a function of the variance of the next layer $\Delta x^{(l+1)}_i$

\begin{align}
    Var\left[ \Delta x_j^{(l)} \right] &= Var\left[ \sum_i W_{ij}^{(l)}\Delta y_i^{(l+1)} \right] \nonumber \\ 
    &= \sum_i Var\left[  W_{ij}^{(l)}\Delta y_i^{(l+1)} \right]  \nonumber \\
    &= n^{(l+1)} Var\left[  W_{ij}^{(l)}\Delta y_i^{(l+1)} \right] \nonumber \\
    &= n^{(l+1)} \left( E\left[W_{ij}^{(l)^2}\right]E\left[\Delta y_{i}^{(l+1)^2}\right] -  E\left[W_{ij}^{(l)}\right]^2E\left[\Delta y_{i}^{(l+1)}\right]^2 \right)  \nonumber \\
    &= n^{(l+1)}  Var\left[W_{ij}^{(l)^2}\right]E\left[\Delta y_{i}^{(l+1)^2}\right] 
   \label{supp_eq:05_02_bckw_ANN_1}
\end{align}

We note that 

\begin{align}
    E\left[ \Delta y_i^{(l+1)} \right] &= E \left[  \Delta x_i^{(l+1)} f'\left(y_i^{(l+1)}  \right)\right] \nonumber \\
    &= E \left[  \Delta x_i^{(l+1)} \right] E\left[f'\left(y_i^{(l+1)}  \right)\right] \nonumber \\
    &= E \left[  \Delta y_i^{(l+2)} \right] E \left[  W_{ij}^{(l+1)} \right] E\left[f'\left(y_i^{(l+1)}  \right)\right] \nonumber \\
    &= 0 \nonumber
\end{align}

Since $E \left[   W_{ij}^{(l+1)} \right] \!=\!0$ and we assumed that $\Delta x_i^{(l+1)}$ is independent of $f'\left(y_i^{(l+1)}\right)$. Then we have 

\begin{align}
    E\left[ \Delta y_i^{(l+1)^2} \right] &= Var \left[  \Delta y_i^{(l+1)}\right] \nonumber \\
    &= Var \left[  \Delta x_i^{(l+1)} f'\left(y_i^{(l+1)}  \right) \right] \nonumber \\
    &= E \left[  \Delta x_i^{(l+1)^2}\right] E\left[ f'\left(y_i^{(l+1)} \right)^2 \right] - E \left[  \Delta x_i^{(l+1)}\right] E\left[ f'\left(y_i^{(l+1)}  \right) \right]  \nonumber \\
    &= Var \left[  \Delta x_i^{(l+1)}\right] E\left[ f'\left(y_i^{(l+1)}  \right)^2 \right]
   \label{suppeq:05_02_bckw_ANN_2}
\end{align}

Finally, the value of $E\left[ f'\left(y_i^{(l+1)} \right)^2 \right]$ depends on the activation function we used.  In the case of a ReLU function its derivative is given by $f'(x>0)=1$ and $f(x\leq 0)=0$. Using the fact that $y_i^{(l+1)}$ is symmetric zero mean as we saw in section \ref{supp_sec:05_02_ann_fwd}.

\begin{align}
   E\left[ f'\left(y_i^{(l+1)} \right)^2 \right] &= \frac{1}{2}\cdot 0+ \frac{1}{2}\cdot 1 = \frac{1}{2} 
   \label{suppeq:05_02_bckw_ANN_3}
\end{align}

Gathering results from \eqref{supp_eq:05_02_bckw_ANN_1}, \eqref{suppeq:05_02_bckw_ANN_2} and \eqref{suppeq:05_02_bckw_ANN_3} we obtaining the final expression

\begin{align}
    Var\left[ \Delta x_j^{(l)} \right] = \frac{1}{2} n^{(l)}  Var\left[W_{ij}^{(l+1)}\right]Var\left[\Delta x_{i}^{(l+1)}\right] 
\end{align}

which for $L$ layers leads to 

\begin{align}
    Var\left[\Delta x_j^{(1)} \right]&= Var\left[ \Delta x_i^{(L-1)} \right] \prod_{l=1}^{L-2} \frac{1}{2} n^{(l+1)}Var\left[  W^{(l)}_{ij} \right]
\end{align}

To avoid exponential variance growth/shrink we set

\begin{align} \label{suppeq:05_02_ann_init_backward}
    Var\left[ W^{(l)}_{ij} \right] = 2/n^{(l+1)}
\end{align}

The expressions found for backward and forward optimal initialisation are different \eqref{supp_eq:05_02_ann_init_forward},\eqref{suppeq:05_02_ann_init_backward}. However, choosing any of them leads to the forward(backward) variance being $1$ while the backward(forward) variance is still a constant \cite{he2015delving}. 

\subsection{Population Firing rate in Spiking Neural Networks} \label{suppsec:05_02_diff_shot}

Computing the firing rate of a spiking neural network is far from trivial due to the one-sided threshold-and-reset mechanism of spiking neurons. Several approaches have been developed to obtain this under different assumptions. 

These include the diffusion approximation \cite{gerstner2014neuronal}, which assumes that the weights are small compared to the distance between the reset and threshold and that weights are restricted to a total of $K$ different values which must include both positive and negative values. We can allow for larger weights as long as they are distributed following a Laplace distribution. This is called shot-noise and generalises the diffusion approximation \cite{richardson2010firing}. Both of these methods are restricted to LIF neurons. For other neuron models the threshold integration method can be used \cite{richardson2007firing, richardson2008spike}.

In addition to the particular assumption to each method, all of them assume a homogeneous (identical neurons) population of neurons and the input spikes being generated by a finite number of homogeneous Poisson processes.

\subsubsection{Diffusion Approximation}


We start with a population of LIF neurons which receive stochastic spikes which are generated following a Poisson process.

\begin{equation} \label{suppeq:05_02_lif}
\frac{dv_j}{dt} = \frac{1}{\tau}(-v_j+ RI_{ext}) + \sum_i \sum\limits_{t_i^{(f)}}  w_{ij}\delta(t-t_i^{(f)}), \qquad   v_j<V_{th}
\end{equation}

We will drop the subscript $j$ and use $v$ to refer to the membrane potential of an arbitrary neuron in the population. Note that $v$ is a random variable.

We are first interested in finding the probability density of the membrane potential of the population of neurons. Since the probability density may vary over time, we define  $P(v, t)$ as the probability that a neuron in the unconnected population has membrane potential $v$ at time $t$. We also define the \textit{flux} $J(v, t)$ as the net fraction of neurons in this population per unit time whose membrane potential crosses $v$ from below.

Neurons in this population receive an external current $I_{ext}$ and stochastic spikes at rate $r_{in}(t)$ and weighted with some weights drawn from a distribution $w_{ij} \! \sim \!p_W(w)$.

If we consider a small voltage interval $[v, v+\delta v]$, the proportion of neurons in this interval at time $t$ is $\int_v^{v+\delta v}P(v', t)dv'$. Consider now that this proportion changes over time. This changes must be equal to the net flux in the interval. In other words, the proportion of neurons that enter the interval per unit time is equal to the net flux of neurons at the extremes of the interval (those that enter from below minus the ones that leave from above). This is expressed as

\begin{align}
    \frac{\partial}{\partial t}\int_v^{v+\delta v} P(v', t)dv' = J(v, t) - J(v+\delta v, t)
\end{align}

Now taking the limit $\delta v \to 0$ and rearranging we obtain the continuity equation

\begin{align}
    \frac{\partial P(v, t)}{\partial t}+ \frac{\partial J(v, t)}{\partial v} = 0
    \label{suppeq:05_02_continuity}
\end{align}

We can also add the resetting and spiking mechanism to the continuity equation by considering that trajectories that reach the threshold $V_{th}$ appear at $V_r$ instantly and this happens at a rate $r_{out}(t)$. Note that $r_{out}(t)=J(V_{th}, t)$.

\begin{align}
    \frac{\partial P(v, t)}{\partial t}+ \frac{\partial J(v, t)}{\partial v} = r_{out}(t) \left(\delta (v-V_r) - \delta (v-V_{th})\right)
    \label{suppeq:05_02_continuity_reset}
\end{align}

We can obtain an expression for the flux as the sum of two contributions, the first one due to the incoming stochastic spiking and the second one due to the deterministic external current and neural dynamics

\begin{align}
    J(v, t)\!=\! J_{jump}(v, t)\!+\!J_{drift}(v, t)
    \label{suppeq:05_02_flux}
\end{align}

Let's start by computing $J_{jump}(v, t)$. We are interested in the total contribution to the flux created by neurons starting at $u\!<\!v$ and jumping above $v$ after receiving a spike. There are a total of $r_{in}(t)$ spikes per unit time each of them contributing $w\!\sim\!p_W(w)$, thus only weights $w\geq v\!-\!u$ will make neurons at $u$ cross $v$. This is given by $r_{in}(t)\int_{v-u}^{\infty}p_W(w)dw$. We repeat this argument for all $u<v$, leading to

\begin{align}
    J_{jump}(v, t) = r_{in}(t) \int_{-\infty}^{v} P(u, t) \int_{v-u}^{\infty}p_W(w)dw du
    \label{suppeq:05_02_flux_jump}
\end{align}

where we have also weighted the contribution to the flux according to the original density $P(u, t)$. 

The flux $J_{drift}(v, t)$ can be computed by simply considering the \textit{velocity} $\frac{dv}{dt}$ times the proportion of neurons at this potential

\begin{align}
    J_{drift}(v, t) &= \frac{dv}{dt} P(v, t) \nonumber \\ 
    &= \left[ -\frac{v}{\tau} + RI_{ext}\right] P(v, t)
    \label{suppeq:05_02_flux_drift}
\end{align}

where the second equality is obtained from \eqref{suppeq:05_02_lif} when not considering the random input spikes.

We now consider the particular case in which we have $K$ synaptic types giving a probability distribution

\begin{align}
    p_W(w) = \sum_k \Delta_k \delta(w-w_k)
    \label{suppeq:05_02_prob_weight_discrete}
\end{align}

such that $\Delta_k\!>\!0$ is the proportion of neurons of a given type $k$ and $\sum_k\Delta_k\!=\!1$. Each synaptic type contributes a total number of spikes per second given by $r_k(t)=r_{in, k}(t)\Delta_k$. Substituting in \eqref{suppeq:05_02_flux_jump} results in

\begin{align}
    J_{jump}(v, t) &= \int_{-\infty}^{v} P(u, t) \int_{v-u}^{\infty}\sum_k r_k(t) \delta(w-w_k)dw du \nonumber \\
    &= \sum_k r_k(t) \int_{-\infty}^{v} P(u, t) \int_{v-u}^{\infty} \delta(w-w_k)dw du \nonumber \\
    &= \sum_k r_k(t) \int_{v-w_k}^{v} P(u, t)du
    \label{suppeq:05_02_flux_jump_diff}
\end{align}

Where the last equality comes from the fact that $\int_{v-u}^{\infty} \delta (w-w_k)dw$ is $1$ if $v-u<w_k$ and $0$ otherwise. Thus, only when $u>v-w_k$ the outer integral will be non-zero.

Substituting \eqref{suppeq:05_02_flux_jump_diff} and \eqref{suppeq:05_02_flux_drift} into the continuity equation with reset \eqref{suppeq:05_02_continuity_reset} we obtain

\begin{align}
    \frac{\partial P(v, t)}{\partial t}= &-\frac{1}{\tau}\frac{\partial}{\partial v}\left(\left[\phantom{\frac{}{}} -v +RI_{ext} \right] P(v, t)\right) \nonumber \\
    &+ \sum_k r_k(t)\left(P(v-w_k, t)-P(v, t)\right)  \nonumber \\
    &+ r_{out}(t) \left(\delta (v-V_r) - \delta (v-V_{th})\right)
\end{align}

Where we can identify the first term with the drift, the second with the stochastic spikes and the last one with the reset. By expanding $P(v-w_k, t)$ into a Taylor series up to a second order around $v$ we obtain

\begin{align}
    \tau \frac{\partial P(v, t)}{\partial t}= &-\frac{\partial}{\partial v}\left(\left[ -v +RI_{ext} + \tau \sum_k r_k(t) w_k\right] P(v, t)\right) \nonumber \\
    &+ \frac{1}{2} \left[ \tau \sum_k r_k(t) w_k^2\right] \frac{\partial^2}{\partial v^2}P(v, t) \nonumber \\
    &+ r_{out}(t) \left(\delta (v-V_r) - \delta (v-V_{th})\right)
    \label{suppeq:05_02_diffussion_eq}
\end{align}

This is the Fokker-Plank equation of the LIF model. We define the total drive $\mu(t)$ and diffusive noise $\sigma(t)$ to be

\begin{align}
   \mu(t)&= RI_{ext} + \tau \sum_k r_k(t) w_k \\
   \sigma^2(t)&= \tau \sum_k r_k(t) w_k^2
\end{align}

leading to a more compact form

\begin{align}
    \tau \frac{\partial P(v, t)}{\partial t}= &-\frac{\partial}{\partial v}\left(\left[ -v + \mu(t)\right] P(v, t)\right)
    + \frac{\sigma^2(t)}{2}  \frac{\partial^2}{\partial v^2}P(v, t) \nonumber \\
    &+ r_{out}(t) \left(\delta (v-V_r) - \delta (v-V_{th})\right)
    \label{suppeq:05_02_diffussion_eq_compact}
\end{align}

It can be shown that under the diffusion approximation and given constant input rates, the original LIF equation with noisy input spikes \eqref{suppeq:05_02_lif} is equivalent to an Ornstein-Uhlenbeck process given by

\begin{equation}
\tau \frac{dv(t)}{dt} = -v(t)+ \mu +\xi (t)  \qquad   v(t)<V_{th}
\label{suppeq:05_02_LIF_OU}
\end{equation}

where $\xi(t) \sim \mathcal{N}\left(0, \sigma^2\tau \right)$ is uncorrelated white Gaussian noise. This leads to the steady state membrane potential mean and variance

\begin{align} 
   \mu_v &= RI_{ext} + \tau \sum_k r_k w_k \label{suppeq:05_02_steady_state_params} \\
   \sigma_v^2 &= \frac{\tau}{2} \sum_k r_k w_k^2
   \label{suppeq:05_02_steady_state_params2}
\end{align}


Without a threshold, the membrane distribution should be Gaussian at all times with time varying mean and variance until it reaches the steady-state distribution with parameters given by \eqref{suppeq:05_02_steady_state_params} and \eqref{suppeq:05_02_steady_state_params2}. However, adding the threshold and reset changes this distribution as it is equivalent to adding a boundary condition onto the Fokker-Planck equation \eqref{suppeq:05_02_diffussion_eq_compact}.

The diffusion equation \eqref{suppeq:05_02_diffussion_eq_compact} does not have closed form solution in general, however, assuming that the input rates are constant $r_k(t)=r_k$, we can obtain the stationary (or steady-state) distribution. That is, the time independent distribution of membrane potential $P(v)$ which is eventually reached when $\frac{\partial P(v, t)}{\partial t}\!=\!0$. 

Using the fact that $P(v, t)$ is a probability distribution and as such it must add up to $1$ and that the density above the threshold should be $0$, it can be shown that the stationary distribution is given by

\begin{align}
P(v)=
    \begin{dcases}
        r_{out}\frac{2\tau }{\sigma^2} \exp \left(-\frac{(v-\mu)^2}{\sigma^2} \right)  \int_{V_r}^{V_{th}}\exp  \left(\frac{(v-\mu)^2}{\sigma^2} \right)  \qquad v\leq V_r\\
        r_{out}\frac{2\tau }{\sigma^2} \exp \left(-\frac{(v-\mu)^2}{\sigma^2} \right)  \int_{v}^{V_{th}}\exp \left(\frac{(v-\mu)^2}{\sigma^2} \right)  \qquad V_r < v \leq V_{th}\\ 
    \end{dcases}
    \label{suppeq:05_02_diff_dist}
\end{align}

where the output firing rate $r_{out}$ is obtained by solving Siegert's integral

\begin{align}
\frac{1}{r_{out}}= \tau \sqrt{\pi} \int_{\frac{V_r-\mu}{\sigma}}^{\frac{V_{th}-\mu}{\sigma}} \exp{\left( x^2\right)}\left( 1 + \text{erf}(x) \right) dx\
\label{suppeq:05_02_siegert}
\end{align}

where $\text{erf}(x)\!=\!\frac{2}{\sqrt{\pi}}\int_0^x\exp(-{x'}^2)dx'$. 

In brief, the diffusion approximation allows as to compute the expected output firing rate of a LIF neuron which receives independent stochastic input spikes through several synaptic types provided the rates are constant and the synaptic weights are small.

As a final note, these results can be extended for different weight distributions beyond simply having $K$ synaptic types. Moreover, it is also possible to extend the diffusion approximation for other synaptic kernels beyond the delta synapses used here. In the end, the diffusion approximation only requires the mean and variance of the membrane potential. 

\subsubsection{Shot-noise}

The main disadvantage of the diffusion approximation introduced before is that it requires the weights to be small. In cases where this condition is not met, the theoretical output firing rate obtained using \eqref{suppeq:05_02_siegert} will not properly reflect the empirical firing rate. Similarly the stationary membrane distribution in \eqref{suppeq:05_02_diff_dist} will also be inaccurate.

The small weights assumption can be dropped if we consider that synaptic weights follow an exponential distribution. Namely, considering both positive and negative weights distributions

\begin{align}
    P_{W_e}(w) &= \frac{1}{w_e}\exp\left(-\frac{w}{w_e}\right), \qquad w>0 \label{suppeq:05_02_shot_dist_e} \\
    P_{W_i}(w) &= \frac{1}{w_i}\exp\left(-\frac{w}{w_i}\right), \qquad w<0
    \label{suppeq:05_02_shot_dist_i}
\end{align}

with $w_e\!>\!0$ and $w_i\!<\!0$. Note that if $w_e=-w_i$ we can describe both positive and negative weights with a Laplace distribution.

The total flux can be written as a combination of the drift contribution and the jump contribution of each type excitatory and inhibitory separately.

\begin{align}
    J(v, t) = -\frac{v}{\tau}P(v, t) + J_e(v, t) + J_i(v, t)
    \label{suppeq:05_02_flux_shot}
\end{align}

where we assumed the external current is zero.

Substituting  the synaptic distributions \eqref{suppeq:05_02_shot_dist_e} and \eqref{suppeq:05_02_shot_dist_i} into the flux expression \eqref{suppeq:05_02_flux_jump} we note that $J_e$ and $J_i$ are solutions to the following differential equations respectively

\begin{align}
    \frac{\partial J_e(v, t)}{\partial v} &= -\frac{J_e(v, t)}{w_e}+r_e P(v, t) - r_{out}(t)\delta(v-V_{th}) \nonumber \\
    \frac{\partial J_i(v, t)}{\partial v} &= -\frac{J_i(v, t)}{w_i}+r_i P(v, t)
    \label{suppeq:05_02_flux_shot_diffeq}
\end{align}

where $r_e$ and $r_i$ are the steady-state excitatory and inhibitory firing rates respectively. This is a direct consequence of choosing the weights to be exponentially distributed. Note that we have added $r_{out}(t)$ in the excitatory term to account for the threshold.

After applying the bilateral Laplace transform on equations \eqref{suppeq:05_02_continuity_reset} and \eqref{suppeq:05_02_flux_shot_diffeq} it is possible to obtain the following expression for the stationary output firing rate

\begin{align}
    \frac{1}{r_{out}} = \tau \int_0^{1/w_e}\frac{Z^{-1}_0(x)}{x} \left( \frac{e^{xV_{th}}}{1-xw_e}-e^{xV_r} \right) dx
    \label{suppeq:05_02_shot_noise_rate}
\end{align}

with $Z_0^{-1}(x) = (1-xw_e)^{\tau r_e}(1-xw_i)^{\tau r_i}$. 

Moreover, we can also obtain the Laplace transform of the probability distributions. In principle it is possible to obtain compute the membrane potential probability distribution by applying the inverse Laplace transform. However, in practise it is difficult to do this and we resort to other methods to obtain it. Namely, threshold integration which is discussed in the next section.

\subsection{Threshold integration}


Threshold integration is a different method that allows to numerically find the membrane distribution and firing rate of spiking populations under the diffusion or shot noise case. This section follows the reasoning and derivations in \cite{richardson2007firing}.

Regardless of which neuron model we use, we always have the continuity equation and a Fokker-Planck equation which relates the time derivative of the distribution with derivatives up to a second order in the voltage

\begin{align}
    \frac{\partial P(v, t)}{\partial t} &+ \frac{\partial J(v, t)}{\partial v} = 0 \\    
    \frac{\partial P(v, t)}{\partial t} &= \mathcal{L} (P(v, t))
\end{align}

We note that if we combine and rearrange these equations we get

\begin{align}
    \mathcal{L} (P(v, t)) &= - \frac{\partial J(v, t)}{\partial v}
\end{align}

since the operator $\mathcal{L}$ consists only on derivatives up to a second order in the voltage we can define a new operator $\mathcal{J}$ which contains only derivatives up to a first order in the voltage that relates the distribution and the flux

\begin{align}
    \mathcal{J} (P(v, t)) &= J(v, t)
\end{align}

We have already seen some examples of the $\mathcal{J}$ operator in the drift and jump fluxes \eqref{suppeq:05_02_flux_jump}, \eqref{suppeq:05_02_flux_drift}, the simplified jump flux we used for the diffusion approximation \eqref{suppeq:05_02_flux_jump_diff} and, the shot noise flux \eqref{suppeq:05_02_flux_shot} \eqref{suppeq:05_02_flux_shot_diffeq}.

This allows us to obtain the stationary distribution by simply setting $\frac{\partial P(v, t)}{\partial t}\!=\!0$ in the continuity equation and explicitly adding the boundary conditions imposed by the threshold and reset. This leads to two simultaneous first-order linear differential equations, namely the stationary continuity equation and the relationship between the membrane potential and the flux

\begin{align}
    \frac{\partial J(v)}{\partial v} &= r_{out}\left( \delta(v-V_r)-\delta(v-V_{th})) \right) \label{suppeq:05_02_threshold_int_init1}\\
    \mathcal{J}(P(v)) &= J(v)
    \label{suppeq:05_02_threshold_int_init2}
\end{align}

Note that the solutions are proportional to the steady-state firing rate giving

\begin{align}
    J(v) &= r_{out}j(v) \nonumber \\ 
    P(v) &= r_{out}p(v) 
    \label{suppeq:05_02_threshold_int_sols}
\end{align}

Once $\mathcal{J}$ has been established, it is possible to numerically integrate backwards from $V_{th}$ to some lower bound voltage $V_{lb}$ using initial conditions $j(V_{th})\!=\!1$ and $p(V_{th})\!=\!0$ (note that $J(V_{th})\!=\!r_{out} = r_{out}j(V_{th})$). Then, $r_{out}$ can be recovered as the reciprocal of to integral $\int_{V_{lb}}^{V_{th}} p(v)dv$ since $P(v)$ must integrate to $1$.

For instance, for the diffusion approximation of the LIF model, we can rearrange the Fokker-Plank equation with boundary conditions \eqref{suppeq:05_02_diffussion_eq_compact} to obtain

\begin{align}
    \frac{\partial P(v, t)}{\partial t} + \frac{\partial}{\partial v} \left( \frac{\mu(t)-v}{\tau}P(v, t) - \frac{\sigma^2(t)}{2\tau}  \frac{\partial}{\partial v}P(v, t) \right)
    \nonumber = r_{out}(t) \left(\delta (v-V_r) - \delta (v-V_{th})\right)
\end{align}

Then, identifying terms in \eqref{suppeq:05_02_threshold_int_init1} we get

\begin{align}
    J(v, t) = \frac{\mu(t)-v}{\tau}P(v, t) - \frac{\sigma^2(t)}{2\tau}  \frac{\partial}{\partial v}P(v, t)
\end{align}

and finally in the steady-state we get the operator 

\begin{align}
    J(v) = \mathcal{J}(P(v)) = \left( \frac{\mu-v}{\tau} - \frac{\sigma^2}{2\tau} \frac{\partial}{\partial v} \right) P(v)
    \label{suppeq:05_02_LIF_operator}
\end{align}

and thus we can rearrange to obtain the following simultaneous equations

\begin{align}
    \frac{\partial J(v)}{\partial v} &= r_{out}\left( \delta(v-V_r)-\delta(v-V_{th}) \right) \\
    \frac{\partial P(v)}{\partial v} &= \frac{2\tau}{\sigma^2}\left( \frac{v-\mu}{\tau}P(v) + J(v) \right)
\end{align}

Note how using \eqref{suppeq:05_02_threshold_int_sols} the unknown output firing rate can be factored out resulting in two simultaneous linear differential equations that can be easily solved

\begin{align}
    \frac{\partial j(v)}{\partial v} &= \delta(v-V_r)-\delta(v-V_{th}) \\
    \frac{\partial p(v)}{\partial v} &= \frac{2\tau}{\sigma^2}\left( \frac{v-\mu}{\tau}p(v) + j(v) \right)
\end{align}

Threshold integration can also be applied for the shot-noise case \cite{richardson2010firing}.

\newpage
\section{Gradient of an SNN layer}\label{app:07_grad_snn}

The unrolled expression of an SNN layer is given by

\begin{align}
v^{(l+1)}[t+1] &= \sum_{k=0}^{t} \alpha^{t-k} \left( W^{(l)}s^{(l)}[k] + (1-\alpha) RI_{ext}\right) \\
s^{(l+1)}_j[t+1] &= H(v^{(l+1)}_j[t+1]-V_{th})
\label{eq:07_grad_H(v)}
\end{align}

where $v^{(l+1)}[t+1], s^{(l+1)}[t+1] \in \mathbb{R}^{n^{(l+1)}}$, $W^{(l)}\in \mathbb{R}^{{n^{(l+1)}}\times {n^{(l)}}}$, $s^{(l)}[t+1] \in \mathbb{R}^{n^{(l)}}$, the rest of parameters are scalars and $H$ is the Heaviside step function.

We assume that the loss function applied to the last layer $\varepsilon : \mathbb{R}^{n^{(L-1)}} \to \mathbb{R}$ is a function of the sum of potentials in the last layer across time (i.e.  $\varepsilon(v^{(L-1)}[t])\!=\!\mathcal{L}\left(\sum_t v^{(L-1)}[t]\right)$). Thus, we can write

\begin{align}
    \frac{\partial \varepsilon}{\partial s^{(l)}[t]} &= \sum_{k>t} \frac{\partial \varepsilon}{\partial v^{(l+1)}[k]} \frac{\partial v^{(l+1)}[k]}{\partial s^{(l)}[t]}
    \label{eq:07_grad_grad_snn0}
\end{align}

where $\frac{\partial \varepsilon}{\partial s^{(l)}[t]}\in\mathbb{R}^{1\times n^{(l)}}$, $\frac{\partial \varepsilon}{\partial v^{(l+1)}[k]}\in\mathbb{R}^{1\times n^{(l+1)}}$ and $\frac{\partial v^{(l+1)}[k]}{\partial s^{(l)}[t]}\in\mathbb{R}^{n^{(l)}\times n^{(l+1)}}$ (using the numerator layout convention).

\begin{figure}[!ht]
\centering
\includegraphics[width=0.8\textwidth]{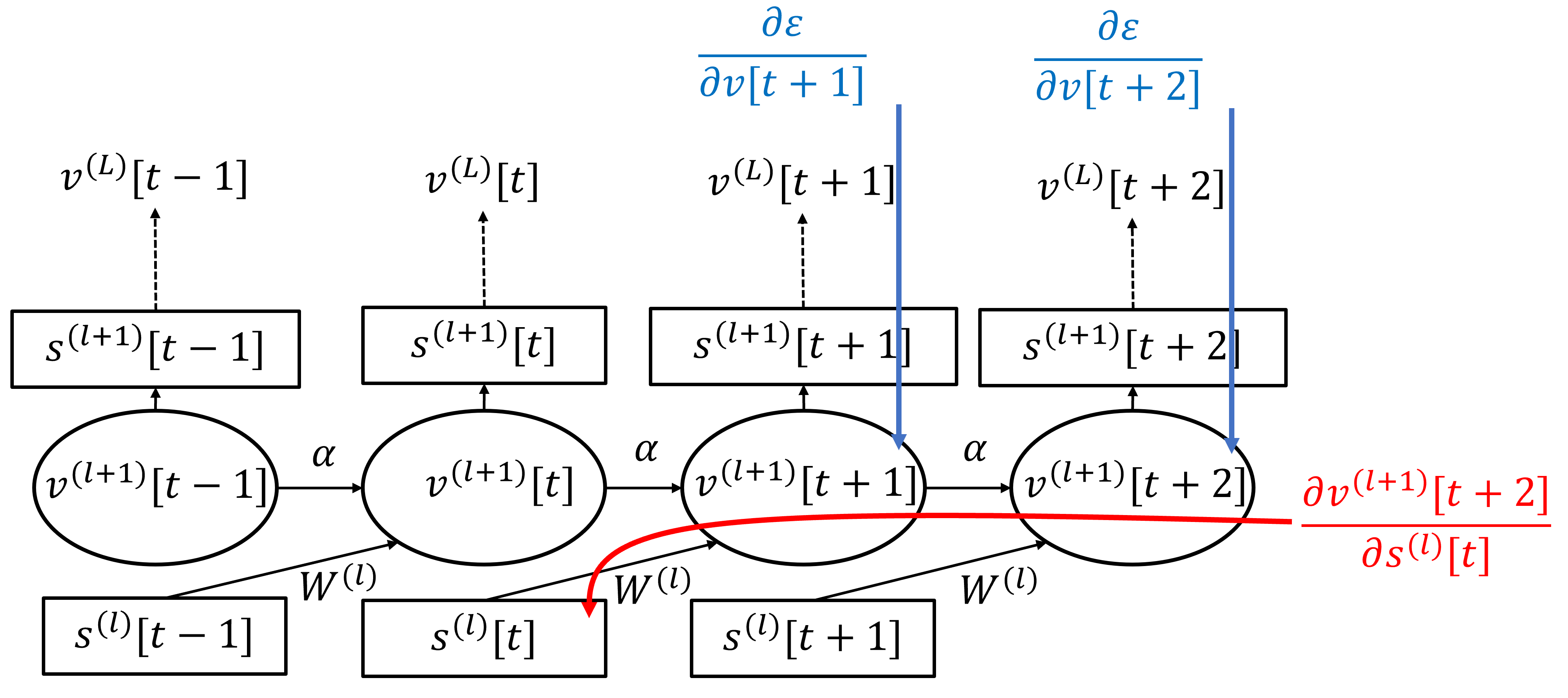}
\caption{Activation and gradient flow diagram for an unrolled SNN}
\label{fig:07_grad_diagram}
\end{figure}

Using equation \eqref{eq:07_grad_H(v)} we can expand the first term  in \eqref{eq:07_grad_grad_snn0}

\begin{align}
    \left( \frac{\partial \varepsilon}{\partial v^{(l+1)}[t]} \right)_i &= \left( \frac{\partial \varepsilon}{\partial s^{(l+1)}[t]} \right)_i H'\left(v_i^{(l+1)}[t]\right)
    \label{eq:07_grad_grad_snn1}
\end{align}

Where $H'$ is a function that subtracts the threshold $V_{th}$ and then applies the \textit{surrogate derivative} of the Heaviside function.

For the second term, we note that a spike at time $t$ can only affect spikes in the future (check Figure \ref{fig:07_grad_diagram} for a visualisation), thus

\begin{align}
\frac{\partial v^{(l+1)}[k]}{\partial s^{(l)}[t]} = 
\begin{cases}
    \frac{\partial v^{(l+1)}[t+1]}{\partial s^{(l)}[t]}\alpha^{k-t-1}, \qquad &k>t \\
    0, \qquad  &\text{otherwise}
\end{cases}
\label{eq:07_grad_grad_snn2}
\end{align}

we also have $\frac{\partial v^{(l+1)}[t+1]}{\partial s^{(l)}[t]}=\frac{\partial W^{(l)}s^{(l)}[t]}{\partial s^{(l)}[t]}=W^{(l)}$.

For notational clarity, we often use $\frac{\partial \varepsilon}{\partial z}:= \Delta z$. Thus, after applying equations \eqref{eq:07_grad_grad_snn2}, equations \eqref{eq:07_grad_grad_snn0} and \eqref{eq:07_grad_grad_snn1} become

\begin{align}
    \Delta s^{(l)}[t] &= \sum_{k>t} \alpha^{k-t-1} \Delta v^{(l+1)}[k] W^{(l)}  \\
    \Delta v_i^{(l+1)}[t] &= \Delta s_i^{(l+1)}[t] H'\left(v_i^{(l+1)}[t]\right)
    \label{eq:07_grad_SNN_gradient}
\end{align}

\newpage
\section{Supplementary results on Firing rate collapse correction} \label{app:07_supp_ch5}

\begin{figure}[!ht]
\centering
\includegraphics[width=0.8\textwidth]{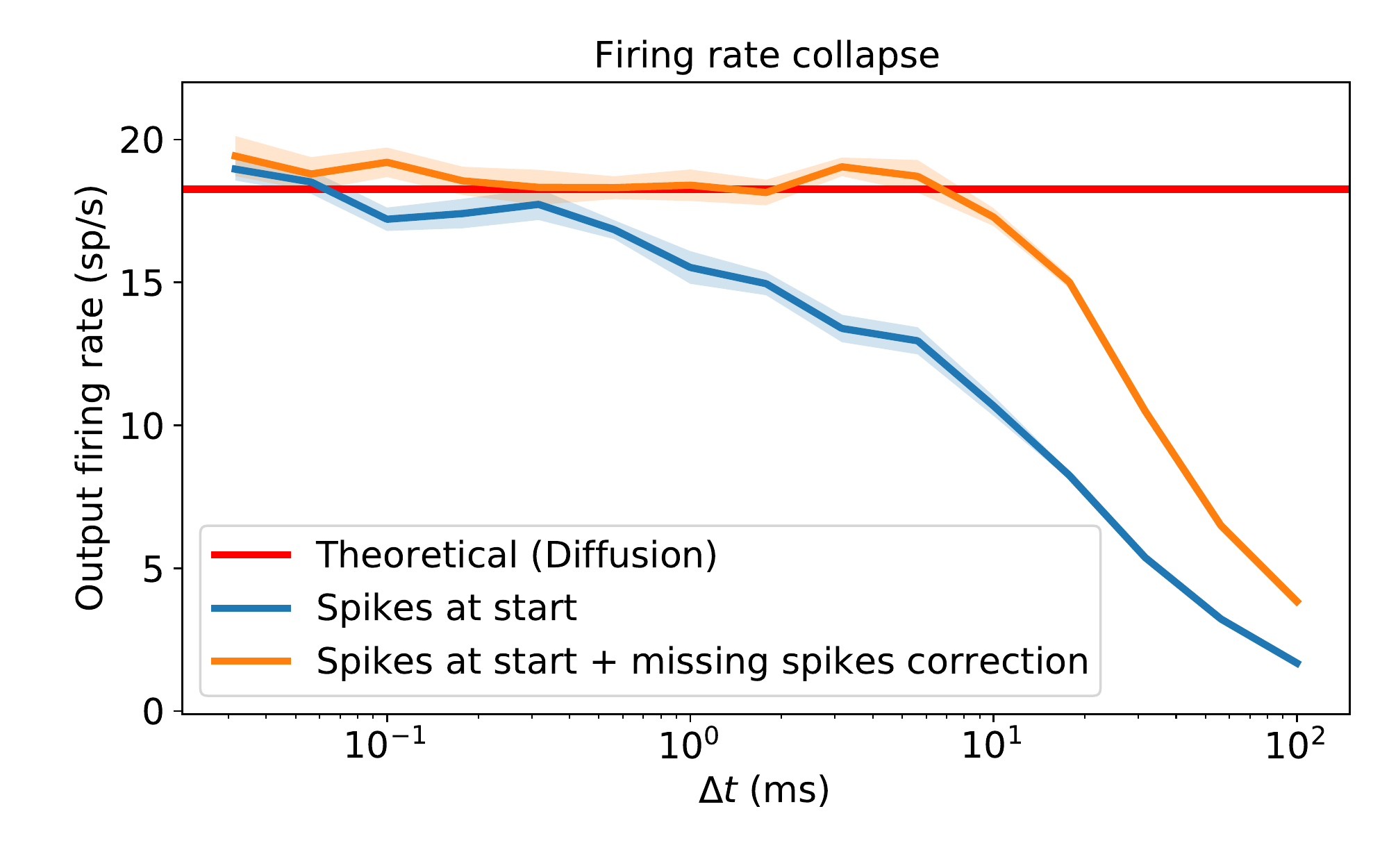}
\caption{Firing rate collapse as the simulation timestep increases and correction. Simulation details are identical to Figure \ref{fig:05_03_collapse}. The integration is performed identically but neurons that did not spike but could have spiked are made to spike randomly according to Theorem \ref{th:05_03_random_walk_correct} but now we use $\bar{\alpha}$ for the computation of $y$. Failure after $\Delta t\!=\!20ms$ is due to the $50$Hz Poisson generators no longer able to keep up.}
\label{fig:07_03_random_walk_correction}
\end{figure}
\newpage
\section{Proofs on Vanishing-Exploding gradients on SNNs} \label{app:07_proofs_ch5}

\thFwd*

\begin{proof*}
\phantom{.}

We want to compute $Var\left[ v_j^{(l+1)}[t+1] \right]$ as a function of parameters of the previous layer. We begin with

\vspace{-0.6cm}
\begin{align} \label{eq:05_04_proof1}
    Var\left[ v_j^{(l+1)}[t+1] \right] &= Var\left[ \sum_{k=0}^{t} \alpha^{t-k} \left( \sum_i w^{(l)}_{ij}s^{(l)}_i[k] + (1-\alpha) RI_{ext}\right) \right] \nonumber \\
    &= Var\left[ \sum_{k=0}^{t} \alpha^{t-k} \left( \sum_i w^{(l)}_{ij}s^{(l)}_i[k] \right) \right] \nonumber \\
    &= Var\left[ \sum_i w^{(l)}_{ij} \sum_{k=0}^{t} \alpha^{t-k}  s^{(l)}_i[k] \right] \nonumber \\
    &= Var\left[ \sum_i w^{(l)}_{ij} c^{(l)}_i[t] \right] \nonumber \\
    &= n^{(l)}Var\left[ w^{(l)}_{ij} \right] E\left[ \left( c^{(l)}_i[t] \right)^2 \right] 
\end{align}

Since we assumed spikes times are independent (Assumption \ref{as:05_04_fw4}) we have that $Cov \left[ s_i^{(l)}[k], s_i^{(l)}[p]  \right]\!=\! Var \left[ s_i^{(l)}[k]  \right]\delta_{pk}$ and we can expand 

\vspace{-0.4cm}
\begin{align} \label{eq:05_04_proof2}
    E\left[ \left( c^{(l)}_i[t] \right)^2 \right] &= E\left[ \left( \sum_{k=0}^t \alpha^{t-k} s^{(l)}_i[k] \right)^2 \right] \nonumber \\
    &= E\left[ \sum_{k=0}^t \sum_{p=0}^t \alpha^{t-k}\alpha^{t-p} s^{(l)}_i[k]s^{(l)}_i[p]  \right] \nonumber \\
    &= \sum_{k=0}^t \sum_{p=0}^t \alpha^{t-k}\alpha^{t-p} E\left[ s^{(l)}_i[k]s^{(l)}_i[p]  \right] \nonumber \\
    &= \sum_{k=0}^t \sum_{p=0}^t \alpha^{t-k}\alpha^{t-p} Cov\left[ s^{(l)}_i[k], s^{(l)}_i[p]  \right] +
    \sum_{k=0}^t \sum_{p=0}^t \alpha^{t-k}\alpha^{t-p} E\left[ s^{(l)}_i[k]  \right] E\left[ s^{(l)}_i[p]  \right]    \nonumber \\
    &= \sum_{k=0}^t \left(\alpha^{t-k}\right)^2 Var\left[ s^{(l)}_i[k] \right] + \left( \sum_{k=0}^t \alpha^{t-k} E\left[ s^{(l)}_i[k]  \right]\right)^2
\end{align}

Then, as we assumed spike times are independent and spikes occur with probability $\rho^{(l)}$ (Assumption \ref{as:05_04_fw5}) we have that $E\left[ s^q  \right]=\int (1-\rho^{(l)})s^q\delta(s)ds + \int \rho^{(l)}s^q\delta(s-1)ds = \rho^{(l)} $. Substituting in \eqref{eq:05_04_proof2}

\begin{align} \label{eq:05_04_proof3}
    E\left[ \left( c^{(l)}_i[t] \right)^2 \right] &= \left( \rho^{(l)}- \left(\rho^{(l)}\right)^2\right) \sum_{k=0}^t \left(\alpha^{t-k}\right)^2 + \left(\rho^{(l)}\right)^2 \left( \sum_{k=0}^t \alpha^{t-k} \right)^2 \nonumber \\
    &= \left( \rho^{(l)}- \left(\rho^{(l)}\right)^2\right) \frac{1-\left(\alpha^2\right)^{t+1}}{1-\alpha^2} + \left(\rho^{(l)}\right)^2 \left( \frac{1-\alpha^{t+1}}{1-\alpha} \right)^2 \nonumber \\
    &\approx \left( \rho^{(l)}- \left(\rho^{(l)}\right)^2\right) \frac{1}{1-\alpha^2} + \left(\rho^{(l)}\right)^2 \left( \frac{1}{1-\alpha} \right)^2
\end{align}

where the last approximation holds for large enough $t$. For instance, if $\Delta t\!=\!1ms$ and $\tau\!=\!10ms$ after 50 timesteps the relative error of the approximation is lower than $1.5\%$ for both sums.

With this result we obtain a final expression by substituting \eqref{eq:05_04_proof3} back in \eqref{eq:05_04_proof1}

\begin{align} \label{eq:05_04_proof4}
    Var\left[ v_j^{(l+1)}[t+1] \right] &\approx n^{(l)}Var\left[ w^{(l)}_{ij} \right] \left( \left( \rho^{(l)}- \left(\rho^{(l)}\right)^2\right) \frac{1}{1-\alpha^2} + \left(\rho^{(l)}\right)^2 \left( \frac{1}{1-\alpha} \right)^2 \right) \nonumber \\
    &= n^{(l)}Var\left[ w^{(l)}_{ij} \right] \left( \frac{1}{1-\alpha^2}\rho^{(l)} + \frac{2\alpha}{(1-\alpha)^2(1+\alpha)}\left(\rho^{(l)}\right)^2\right) \nonumber \\
    &= n^{(l)}Var\left[ w^{(l)}_{ij} \right] \left( A(\alpha)\rho^{(l)} + B(\alpha)\left(\rho^{(l)}\right)^2\right)
\end{align}

Using Chebyshev's inequality and the fact that we assumed the membrane potential is symmetric around zero we can write

\begin{align}
    P(|v_j^{(l+1)}[t+1]|\geq V_{th})&= 2P(v_j^{(l+1)}[t+1]\geq V_{th}) \nonumber \\
    &<\frac{ Var\left[ v_j^{(l+1)}[t+1] \right]}{V_{th}^2}
\end{align}

Since $\rho^{(l+1)}=P(v_j^{(l+1)}[t+1]\geq V_{th})$ we have

\begin{align}
    \rho^{(l+1)}<\frac{ Var\left[ v_j^{(l+1)}[t+1] \right]}{2V_{th}^2}
\end{align}

Using this on \eqref{eq:05_04_proof4} we finally obtain

\begin{align}
    \rho^{(l+1)}< \frac{n^{(l)}Var\left[ w^{(l)}_{ij} \right]}{2V_{th}^2} \left( A(\alpha)\rho^{(l)} + B(\alpha)\left(\rho^{(l)}\right)^2\right)
    \label{eq:forward_ineq}
\end{align}
\raggedleft{\qedsymbol{}}
\end{proof*}

\thBwd*

\begin{proof*}

\phantom{.}

We begin by expanding using Assumptions \ref{as:05_04_bw1} and \ref{as:05_04_bw2}

\begin{align} \label{eq:05_04_proof_bwd_1}
    Var\left[\Delta s_j^{(l)}[t]\right] 
    &= Var\left[ \sum_i w_{ij}^{(l)}\sum_{k>t}\alpha^{k-t-1}\Delta v_i^{(l+1)}[k] \right]  \nonumber\\ 
    &= n^{(l+1)} Var\left[ w_{ij}^{(l)} \sum_{k>t}\alpha^{k-t-1}\Delta v_i^{(l+1)}[k] \right]  \nonumber\\ 
    &= n^{(l+1)} E\left[w_{ij}^{(l)^2}\right]E\left[\left(\sum_{k>t}\alpha^{k-t-1}\Delta v_i^{(l+1)}[k]\right)^2\right]
\end{align}

We further expand the last term using Assumptions \ref{as:05_04_bw3} and \ref{as:05_04_bw4}

\begin{align} \label{eq:05_04_proof_bwd_2}
    E\left[\left(\sum_{k>t}\alpha^{k-t-1}\Delta v_i^{(l+1)}[k]\right)^2\right] 
    &= E\left[\sum_{k>t}\sum_{p>t}\alpha^{k-t-1}\alpha^{p-t-1}\Delta v_i^{(l+1)}[k]\Delta v_i^{(l+1)}[p]\right] \nonumber\\
    &= \sum_{k>t}\sum_{p>t}\alpha^{k-t-1}\alpha^{p-t-1}E\left[\Delta v_i^{(l+1)}[k]\Delta v_i^{(l+1)}[p]\right] \nonumber\\
    &= \sum_{k>t}\sum_{p>t}\alpha^{k-t-1}\alpha^{p-t-1}Cov\left[\Delta v_i^{(l+1)}[k],\Delta v_i^{(l+1)}[p]\right] \nonumber\\ &\quad + \left(\sum_{k>t}\alpha^{k-t-1}E\left[\Delta v_i^{(l+1)}[p]\right]\right)^2 \nonumber \\
    &= \sum_{k>t}\left(\alpha^{k-t-1}\right)^2 Var\left[\Delta v_i^{(l+1)}[k]\right] + \left(\sum_{k>t}\alpha^{k-t-1}E\left[\Delta v_i^{(l+1)}[p]\right]\right)^2 
\end{align}

Substituting \eqref{eq:05_04_proof_bwd_2} back into \eqref{eq:05_04_proof_bwd_1} 

\begin{align} \label{eq:05_04_proof_bwd_3}
    Var\left[\Delta s_j^{(l)}[t]\right] 
    &= n^{(l+1)} E\left[w_{ij}^{(l)^2}\right]\left(\sum_{k>t}\left(\alpha^{k-t-1}\right)^2 Var\left[\Delta v_i^{(l+1)}[k]\right] + \left(\sum_{k>t}\alpha^{k-t-1}E\left[\Delta v_i^{(l+1)}[p]\right]\right)^2\right)   \nonumber \\
    &\approx n^{(l+1)} E\left[w_{ij}^{(l)^2}\right]\left(\frac{1}{1-\alpha^2} Var\left[\Delta v_i^{(l+1)}[k]\right] + \frac{1}{(1-\alpha)^2}\left(E\left[\Delta v_i^{(l+1)}[p]\right]\right)^2\right)   \nonumber \\
\end{align}

where the last equality holds as $k-t$ is increases. 

Now we need to compute the terms $Var\left[\Delta v_i^{(l+1)}[k]\right]$ and $E\left[\Delta v_i^{(l+1)}[p]\right]$. For this, the following result will be useful

\begin{align} \label{eq:05_04_proof_bwd_4}
    E\left[\Delta s_j^{(l+1)}[t]\right] 
    &= E\left[\sum_i w_{ij}^{(l)}\sum_{k>t}\alpha^{k-t-1}\Delta v_i^{(l+1)}[k]\right]\nonumber \\
    &= n^{(l+1)}E\left[w_{ij}^{(l)}\right]E\left[\sum_{k>t}\alpha^{k-t-1}\Delta v_i^{(l+1)}[k]\right]\nonumber \\
    &= 0
\end{align}

Then using Assumption \ref{as:05_04_bw5} we have 

\begin{align}  \label{eq:05_04_proof_bwd_5}
    E\left[\Delta v_i^{(l+1)}[t]\right] 
    &= E\left[\Delta s_i^{(l+1)}[t]H'\left(v_i^{(l+1)}[t]\right)\right]\nonumber \nonumber \\
    &= E\left[\Delta s_i^{(l+1)}[t]\right]E\left[H'\left(v_i^{(l+1)}[t]\right)\right] \\
    &= 0
\end{align}

and also 

\begin{align} \label{eq:05_04_proof_bwd_6}
    Var\left[\Delta v_i^{(l+1)[t]}\right] 
    &= Var\left[\Delta s_i^{(l+1)}[t]H'\left(v_i^{(l+1)}[t]\right)\right]\nonumber \\
    &= E\left[\left(\Delta s_i^{(l+1)}[t]\right)^2\right]E\left[\left(H'\left(v_i^{(l+1)}[t]\right)\right)^2\right] - E\left[\left(\Delta s_i^{(l+1)}[t]\right)\right]^2E\left[H'\left(v_i^{(l+1)}[t]\right)\right]^2 \nonumber \\
    &= E\left[\left(\Delta s_i^{(l+1)}[t]\right)^2\right]E\left[\left(H'\left(v_i^{(l+1)}[t]\right)\right)^2\right]\nonumber \\
    &= E\left[\left(\Delta s_i^{(l+1)}[t]\right)^2\right]\rho_a^{(l+1)}
\end{align}

And finally substituting \eqref{eq:05_04_proof_bwd_5} and \eqref{eq:05_04_proof_bwd_6} into \eqref{eq:05_04_proof_bwd_3} we obtain 

\begin{align} \label{eq:05_04_proof_bwd_7}
    Var\left[\Delta s_j^{(l)}\right] 
    &\approx n^{(l+1)} E\left[w_{ij}^{(l)^2}\right]\left(\frac{1}{1-\alpha^2} Var\left[\Delta v_i^{(l+1)}[k]\right] + \frac{1}{(1-\alpha)^2}\left(E\left[\Delta v_i^{(l+1)}[p]\right]\right)^2\right)   \nonumber \\
    &= n^{(l+1)} E\left[w_{ij}^{(l)^2}\right]\frac{1}{1-\alpha^2} E\left[\left(\Delta s_i^{(l+1)}[t]\right)^2\right]\rho_a^{(l+1)} \nonumber \\
    &= \frac{\rho_a^{(l+1)} }{1-\alpha^2} n^{(l+1)} Var\left[w_{ij}^{(l)}\right] Var\left[\Delta s_i^{(l+1)}[t]\right]
\end{align}
\raggedleft{\qedsymbol{}}
\end{proof*}

\newpage
\section{Network Initialisation limit cases} \label{app:07_init_fail}

\vspace{-0.5cm}
\begin{figure}[!ht]
\centering
\includegraphics[width=0.65\textwidth]{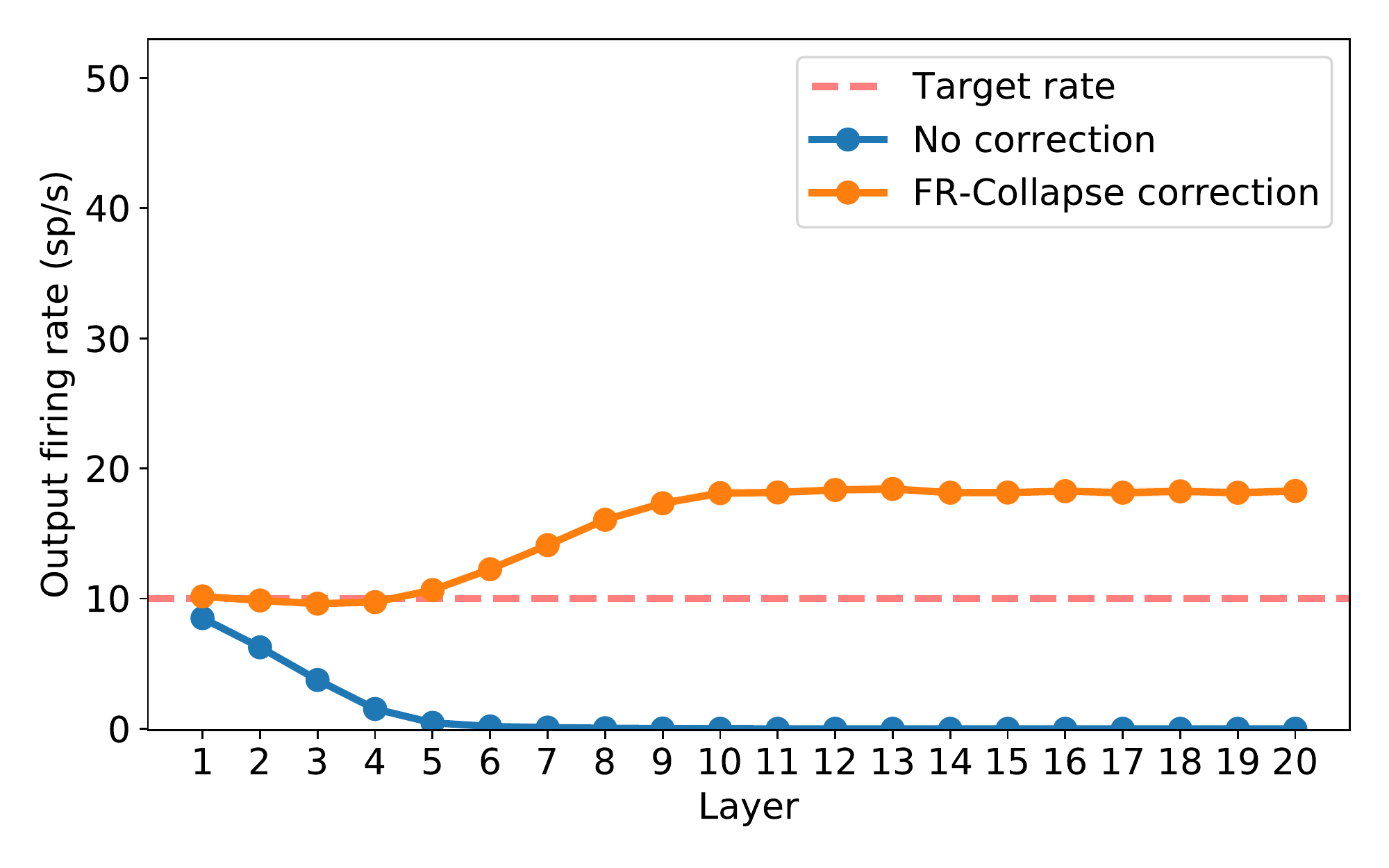}
\vspace{-0.55cm}
\caption{Firing rate at each layer on a feed-forward SNN with 20 layers with target rate $10$Hz. Simulations details are identical to Figure \ref{fig:05_05_forward} but now with a different target rate and input rate ($10$Hz). The optimal weight was $0.030$}
\label{fig:07_forward_fail}
\end{figure}

\begin{figure}[!ht]
\centering
\includegraphics[width=0.65\textwidth]{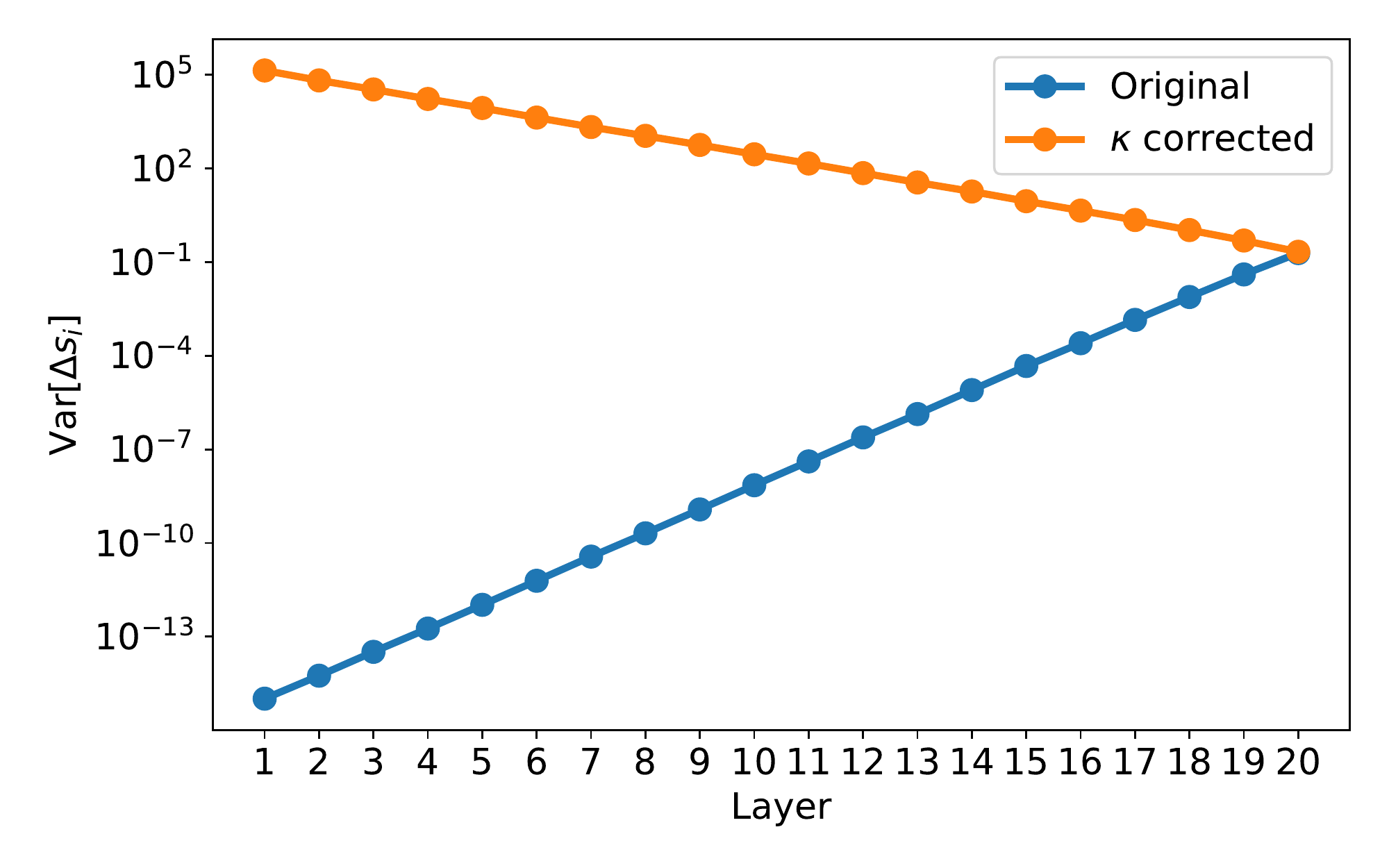}
\vspace{-0.55cm}
\caption{Variance of the spike gradient at each layer on a feed-forward SNN with 20 layers and $B_{th}\!=\!0.1$. Other simulations details are identical to Figure \ref{fig:05_05_backward}.}
\label{fig:07_backward_fail}
\end{figure}

\clearpage

\bibliographystyle{unsrt}
\bibliography{bibfile}

\end{document}